\documentclass{article}
\usepackage{PRIMEarxiv}

\usepackage[utf8]{inputenc} 
\usepackage[T1]{fontenc}    
\usepackage{hyperref}       
\usepackage{url}            
\usepackage{booktabs}       
\usepackage{amsfonts}       
\usepackage{nicefrac}       
\usepackage{microtype}      
\usepackage{lipsum}
\usepackage{fancyhdr}       
\usepackage{graphicx}       
\graphicspath{{media/}}     
\usepackage{amsmath}
\usepackage{amssymb}
\usepackage{physics}
\usepackage{float}
\usepackage{subcaption}
\usepackage[compact]{titlesec}
\usepackage{parskip}

\newtheorem{theorem}{Theorem}
\newtheorem{lemma}{Lemma}

\DeclareMathOperator{\ReLU}{ReLU}
\DeclareMathOperator{\IF}{IF}
\DeclareMathOperator{\ClipFloor}{ClipFloor}
  
\title{Artificial to Spiking Neural Networks Conversion for Scientific Machine Learning
\thanks{\textit{\underline{Citation}}: 
\textbf{Authors. Title. Pages.... DOI:000000/11111.}} 
}

\author{
  Qian Zhang, Chenxi Wu, Adar Kahana, George Em Karniadakis\\
  Division of Applied Mathematics\\
  Brown University \\
  Providence\\
  \texttt{\{qian\_zhang1, chenxi\_wu, adar\_kahana, george\_karniadakis\}@brown.edu} \\
   \And
  Youngeun Kim, Yuhang Li, Priyadarshini Panda \\
  Department of Electrical Engineering \\
  Yale University \\
  New Haven \\
  \texttt{\{youngeun.kim, yuhang.li, priya.panda\}@yale.edu} \\
}

\begin{document}
\maketitle

\newcommand{\e}{\vb{e}}
\newcommand{\ec}{\vb{e}_c}
\newcommand{\Hes}{\vb{H}}
\newcommand{\W}{\vb{W}}
\newcommand{\B}{\vb{B}}
\newcommand{\C}{\vb{C}}
\newcommand{\x}{\vb{x}}
\newcommand{\s}{\vb{s}}
\newcommand{\z}{\vb{z}}

\newcommand{\tf}{\Tilde{f}}
\newcommand{\te}{\Tilde{\e}}
\newcommand{\tHes}{\Tilde{\Hes}}
\newcommand{\tB}{\Tilde{\B}}
\newcommand{\tec}{\Tilde{\ec}}

\newcommand{\px}{\partial\x}
\newcommand{\pz}{\partial\z}
\newcommand{\pL}{\partial L}

\newcommand{\at}[1]{^{(#1)}}
\newcommand{\att}[1]{^{(#1), \top}}

\begin{abstract}
We introduce a method to convert Physics-Informed Neural Networks (PINNs), commonly used in scientific machine learning, to Spiking Neural Networks (SNNs), which are expected to have higher energy efficiency compared to traditional Artificial Neural Networks (ANNs). We first extend  the calibration technique of SNNs to arbitrary activation functions beyond ReLU, making it more versatile, and we prove a theorem that ensures the effectiveness of the calibration. We successfully  convert PINNs to SNNs, enabling computational efficiency for diverse regression tasks in solving multiple differential equations, including the unsteady Navier-Stokes equations. We demonstrate great gains in terms of overall efficiency, including Separable PINNs (SPINNs), which accelerate the training process. Overall, this is the first work of this kind and the proposed method achieves relatively good accuracy with low spike rates. 
\end{abstract}

\section{Introduction}
The use of machine learning techniques in the scientific community has been spreading widely, reaching many fields such as physics \cite{pinn_fluid,ml_physics,JIN2020SympNets, Zhang2022HINTS}, chemistry \cite{ml_chemistry1, ml_chemistry2}, biology \cite{pinn_sysbio,ml_biology, Zhang2022AOSLO}, geophysics \cite{geologics_ccs,geologics_fwi}, epidemiology\cite{Zhang2021PLOS, Kharazmi2021} and many more. The advances in computation capabilities have enabled many researchers to reformulate diverse problems as data-driven problems, by combining prior knowledge of the problem with fitting a model for the available data. A prominent drawback of Scientific Machine Learning (SciML) techniques is that they are usually expensive in terms of computational cost. They require either knowledge of the governing equations that determine the process (approximating them is a costly procedure), or a large amount of data to fit (expensive as well). The SciML community is striving for a more efficient method for training and inferring neural networks. Neuromorphic chips are one edge computing component that SciML applications could benefit from. In this work, we explore methods for enabling this.

An important breakthrough in the field of SciML was the invention of the Physics-Informed Neural Networks (PINNs) \cite{raissi2019physics,lu2021deepxde}. PINNs incorporate the knowledge of the physical experiment and governing equations into the network training step, making it a hybrid (physics and data) training. PINNs and its extensions \cite{pinn_sampling,pinn_attention,g_pinn,Zou2023MHNN} have achieved great success in many fields and applications \cite{pinn_finance,pinn_fluid,pinn_geometry,pinn_hfm,pinn_material,pinn_optics,pinn_sysbio,Zou2022NeuralUQ,PSAROS2023UQ} . A disadvantage of PINNs is that like other deep neural networks, they are prone to long training times. In addition, when changing the problem conditions (initial conditions, boundary conditions, domain properties, etc.), the PINN has to be trained from scratch. Therefore, for real-time applications, a more efficient solution is sought after. For training a PINN, one usually uses smooth activation functions (such as the Tanh or Sine activation functions), where in most ANNs the ReLU activation is dominant. Using smooth activation function in a SNN is a new challenge we address in this paper with theoretical justification.

Spiking Neural Networks (SNNs) have been gaining traction in the machine learning community for the past few years. The main reason is their expected efficiency 
\cite{massa2020efficient,kim2022rate,bouvier2019spiking, roy2019towards}, in terms of energy consumption, compared to their Artificial Neural Network (ANN) counterparts that are commonly used for many applications. In addition, the advances in neuromorphic hardware (such as Intel's Loihi 2 chip \cite{davies2018loihi,orchard2021efficient}), call for innovative algorithms and software that can utilize the chips and produce lighter and faster machine learning models. However, developing an SNN is a challenging task, especially for regression \cite{Kahana2022SpikingNO, Zhang2022SMS}. Studies have been conducted for translating components from the popular ANNs into a spiking framework  \cite{nageswaran2009computing}, but many components are not yet available in the spiking regime. In this paper we focus on that specific aspect.

There are three popular approaches for training SNNs. The first involves using mathematical formulations of the components of the brain, such as the membrane \cite{bienenstock1982theory,hodgkin1952measurement,izhikevich2007dynamical}, the synapse \cite{gerstner2002spiking}, etc. In this case, one uses a Hebbian learning rule \cite{gerstner1993spikes} to find the weights of the synapses (the trainable parameters) using forward propagation (without backward propagation \cite{rumelhart1986learning,lillicrap2020backpropagation}). The second method involves building surrogate models for the elements in the SNN that block the back-propagation, such as the non-differentiable activation functions used in SNNs. The third method, which is discussed in this paper, addresses converting a trained ANN into a SNN.

The main contributions of this paper are as follows:
\begin{enumerate}
    \item We propose a method to convert PINNs, a type of neural network commonly used for regression tasks, to Spiking Neural Networks (SNNs). The conversion allows for utilizing the advantages of SNNs in the inference stage, such as computational efficiency, in regression tasks.
    \item We extend the calibration techniques used in previous studies to arbitrary activation functions, which significantly increases the applicability of the conversion method. Furthermore, we provide a convergence theorem to guarantee the effectiveness of the calibration.
    \item We apply the conversion to separable PINNs (SPINNs), which accelerates the training process of the PINNs.
\end{enumerate} 
Overall, the proposed method extends the application of SNNs in regression tasks and provides a systematic and efficient approach to convert existing neural networks for diverse regression tasks to SNNs.

\section{Related Work}
\paragraph{Physics-informed neural networks (PINNs):}  An innovative framework that combines neural networks with physical laws to learn complex physical phenomena. In PINNs, the physical equations are integrated into the loss function, which allows the network to learn from both the given data and the underlying physics. This approach significantly improves the network's ability to handle incomplete or noisy data and performs well with limited training data. PINNs have been successfully applied to a range of problems in fluid dynamics, solid mechanics, and more \cite{pinn_hfm,pinn_fluid,pinn_geometry,pinn_material,pinn_sysbio,pinn_optics}.
\paragraph{Separable PINNs (SPINNs):} Cho et al. \cite{cho2022separable} proposed a novel neural network architecture called SPINNs, which aims to reduce the computational demands of PINNs and alleviate the curse of dimensionality. Unlike vanilla PINNs that use point-wise processing, SPINN works on a per-axis basis, thereby reducing the number of required network forward passes. SPINN utilizes factorized coordinates and separated sub-networks, where each sub-network takes an independent one-dimensional coordinate as input, and the final output is generated through an outer product and element-wise summation. Because SPINN eliminates the need to query every multidimensional coordinate input pair, it is less affected by the exponential growth of computational and memory costs associated with grid resolution in standard PINNs. Furthermore, SPINNs operate on a per-axis basis, which allows for parallelization with multiple GPUs.
\paragraph{Spiking Neural Networks (SNNs):} A type of Artificial Neural Network (ANN), that differs in the implementation of the core components. The purpose is to create more biologically-plausible training and inference procedures. Unlike traditional ANNs, which process information through numerical values, SNNs process information through spikes, which occur in response to stimulation (much like the human brain). SNNs are becoming increasingly popular as they can mimic the temporal nature of biological neurons. Additionally, SNNs are computationally efficient and have the potential for efficient hardware implementation, making them well-suited for real-time applications. Combining SNN implementation with edge computing, training could be significantly faster, and inference as well \cite{massa2020efficient,kim2022rate,bouvier2019spiking}. Recent results have shown that SNNs can achieve high accuracy on image classification tasks, with up to 99.44\% on the MNIST dataset \cite{rueckauer2017conversion} and up to 79.21\% on ImageNet \cite{li2021free}.
\paragraph{SNN conversion} A technique to transform a trained ANN into a SNN. SNN conversion usually involves mapping the weights and activation functions of the ANN to the synaptic strengths and spike rates of SNNs. It is considered the most efficient way to train deep SNNs, as it avoids the challenges of direct SNN training, such as gradient estimation and spike generation \cite{liu2021optimal,kim2021reducing,snn_toolbox}. The algorithm of SNN conversion can be divided into two steps: offline conversion and online inference. In the offline conversion step, the trained ANN model is converted into an equivalent SNN model by adjusting the network parameters. In the online inference step, the converted SNN model is used for the inference. In the online step, the SNN is intended to be deployed on neuromorphic hardware, to unlock its full potential and energy efficiency.

\section{Method}
A complementary process to the conversion technique. The SNN calibration \cite{li2022calibration} is a method that minimizes the loss of accuracy and efficiency when converting an ANN into a SNN. SNN calibration leverages the knowledge of the pre-trained ANN and corrects the conversion error layer by layer. The algorithm consists of two steps: replacing the ReLU activation function with a spike activation function, and applying the calibration to adjust only the biases (light) or weights and biases (advanced) of each layer. The calibration method is based on the theoretical analysis of the conversion error and its propagation through the network. SNN calibration can achieve comparable or even better performance than the original ANN on various datasets and network architectures. This paper is a generalization of this work. We propose an extension to the SNN conversion, which is apropriate for regression tasks.

\subsection{SNN Conversion setup}
We consider a dataset $D=(X,Y)$, and an ANN $\mathcal{A}$ with $n$ hidden layers, trained to fit it. Let $\x\at{n}=\mathcal{A}(X)$ be the output of $\mathcal{A}$. The goal of SNN conversion is to find an SNN $\mathcal{S}$, whose (averaged) output is $\bar{\s}\at{n}=\mathcal{S}(X)$, such that $\x\at{n}$ is close to $\bar{\s}\at{n}$. In other words, we want to minimize the norm of the error $\e\at{n}\stackrel{d}{=}\x\at{n}-\bar{\s}\at{n}$ for a given $\mathcal{A}$ and $D$. The same network structure as $\mathcal{S}$ is used for $\mathcal{A}$ and the activation function of $\mathcal{A}$ is replaced with IF. Then we can analyze the factors that influence the total conversion error $\e\at{n}$.

\subsection{SNN Conversion with Calibration}
In this section, we will briefly explain the SNN conversion with calibration proposed in \cite{li2022calibration}. Consider an MLP model with $n$ layers, the first $n-1$ layers use ReLU activation and the last layer has no activation. We denote $\W\at{l}$ as the weights and bias for layer $l$.

\begin{figure}[htbp]
    \centering
    \includegraphics[width=0.4\linewidth]{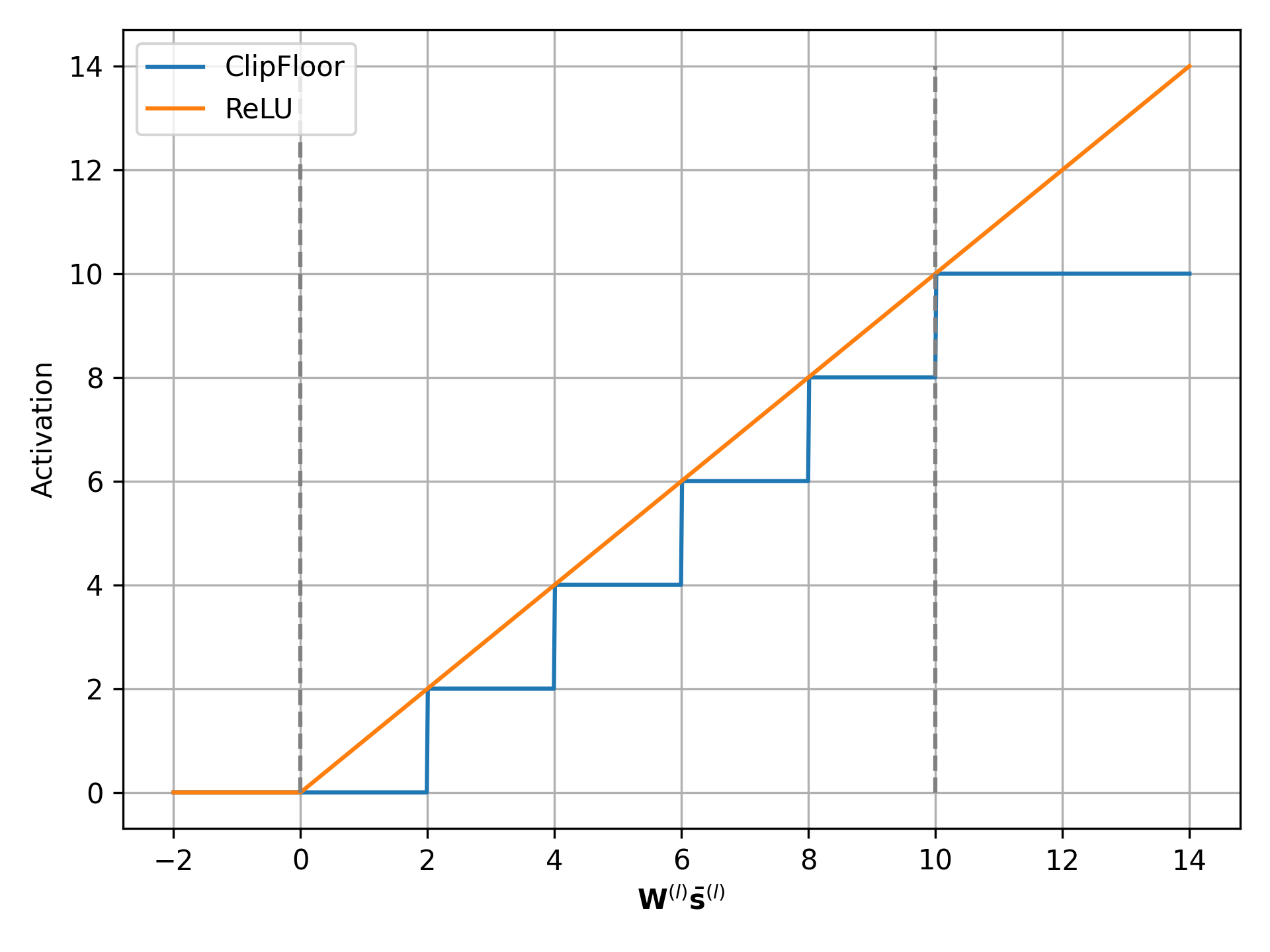}
    \includegraphics[width=0.4\linewidth]{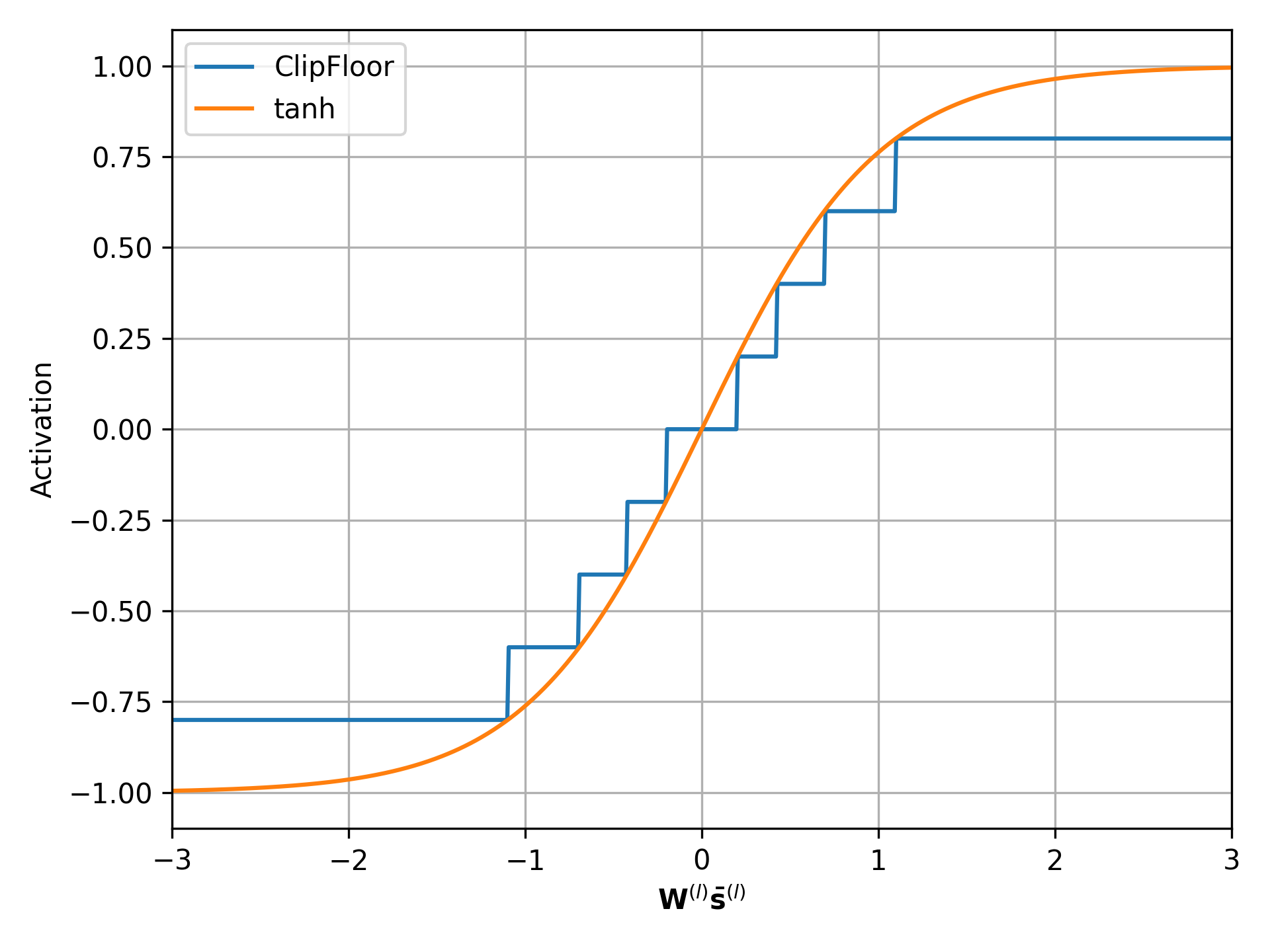}
    \caption{Illustration of $\ClipFloor$ function and its relationship to $\ReLU$ (left) and $\tanh$ (right).}
    \label{fig:method:clipfloor}
\end{figure}

The naive way of SNN conversion is simply replacing the ReLU activation layers with IF activation layers. In this case, we define $\x\at{l}$ as the output of layer $l$ recursively, which is $\x\at{l} = \ReLU(\W\at{l}\x\at{l-1})$ and $\x\at{0}$ is the input. Similarly, we define $\bar{\s}\at{l}$, the output of the layer $l$ of converted SNN, as $\bar{\s}\at{l} = \IF(\W\at{l}\bar{\s}\at{l-1})$ and $\bar{\s}\at{0}=\x\at{0}$ is the input. In fact, we can compute the expected output spikes as $\bar{\s}\at{l} = \ClipFloor(\W\at{l}\bar{\s}\at{l-1})$. The $\ClipFloor$ function is an approximation to $\ReLU$ and is illustrated in Figure \ref{fig:method:clipfloor}.

Then we can define the conversion error of layer $l$ as $\e\at{l} = \x\at{l}-\bar{\s}\at{l}$ and decompose it as
\begin{equation}
\begin{aligned}
    \e\at{l} &= \x\at{l}-\bar{\s}\at{l}\\
    &= \ReLU(\W\at{l}\x\at{l-1})-\ClipFloor(\W\at{l}\bar{\s}\at{l-1})\\
    &= \ReLU(\W\at{l}\x\at{l-1})-\ReLU(\W\at{l}\bar{\s}\at{l-1})+\ReLU(\W\at{l}\bar{\s}\at{l-1})-\ClipFloor(\W\at{l}\bar{\s}\at{l-1})\\
    &= \e\at{l}_r+\e\at{l}_c
\end{aligned}
\label{eqn:method:dissection-relu}
\end{equation}
Here $\e\at{l}_r$ represents the error caused by approximating the continuous input by the spiking input, and $\e\at{l}_c$ represents the local conversion error caused by changing the smooth activation function to the spiking activation (IF). A major result in \cite{li2022calibration} is that we can bound the conversion error $\e\at{n}$, with the weighted sum of the local conversion errors $\e\at{l}_c$. This allows us to minimize the conversion error via optimizing each $\e\at{l}_c$.

\subsection{Results for general activation functions}
In fact, the results can be generalized to activation functions other than $\ReLU$ by similar techniques in \cite{li2022calibration}. Since the generalized activation functions may have negative values, we introduce the idea of negative threshold, a concept in SNNs that allows neurons to fire both positive and negative spikes, depending on their membrane potential \cite{liu2021backeisnn}. A positive spike occurs when the membrane potential exceeds the positive threshold, and a negative spike occurs when it is below the negative threshold. This mimics the biological behavior of neurons that do not fire a spike when the membrane potential does not reach the threshold. Negative threshold can be applied to different types of SNNs and learning functions depending on the problem domain and the data characteristics. It is very helpful when the dataset contains negative values.

To formulate the conversion with calibration for generalized activations, we consider the conversion error decomposition
\begin{equation}
\begin{aligned}
    \e\at{l} &= \x\at{l}-\bar{\s}\at{l}\\
    &= f(\W\at{l}\x\at{l-1})-\ClipFloor(\W\at{l}\bar{\s}\at{l-1})\\
    &= f(\W\at{l}\x\at{l-1})-f(\W\at{l}\bar{\s}\at{l-1})+f(\W\at{l}\bar{\s}\at{l-1})-\ClipFloor(\W\at{l}\bar{\s}\at{l-1})\\
    &= \e\at{l}_r+\e\at{l}_c
\end{aligned}
\end{equation}
where $\e\at{l}_r$ and $\e\at{l}_c$ have the same meaning as in Eq \ref{eqn:method:dissection-relu}. Then we can also use the weighted sum of the local conversion errors $\e\at{l}_c$ to bound the total conversion error $\e\at{n}$ by the following theorem.

\begin{theorem}
For any activation function whose function values and first-order derivatives can be uniformly approximated by piecewise linear functions and up to second-order derivatives are bounded, then the conversion error in the final network output space can be bounded by a weighted sum of local conversion errors, given by
\begin{equation}
    \vb{e}^{(n),\top}\vb{H}^{(n)}e^{(n)} \leq \sum_{l=1}^n 2^{n-l+1} \vb{e}_c^{(l), \top}(\vb{H}^{(l)} + K_L^{(l)}\sqrt{L}\vb{I})\vb{e}_c^{(l)}
\end{equation}
where $L$ is the training loss.\label{thm:main}
\end{theorem}
We present the detailed proof in the Appendix. The main technique is to find piecewise linear functions that approximate the smooth activation function.

Therefore, the total conversion error is bounded from above by the local conversion errors. To achieve more accurate conversion performance, we can minimize the total conversion error by minimizing the local conversion error layerwise, which can be easily implemented as in \cite{li2022calibration}. We enlarge the last layer threshold to preserve the maximum value of the output. The details are discussed in the Appendix.

\section{Results}
\subsection{Function regression}
\begin{figure}[htbp]
    \centering
    \begin{subfigure}{0.3\linewidth}
        \includegraphics[width=\linewidth]{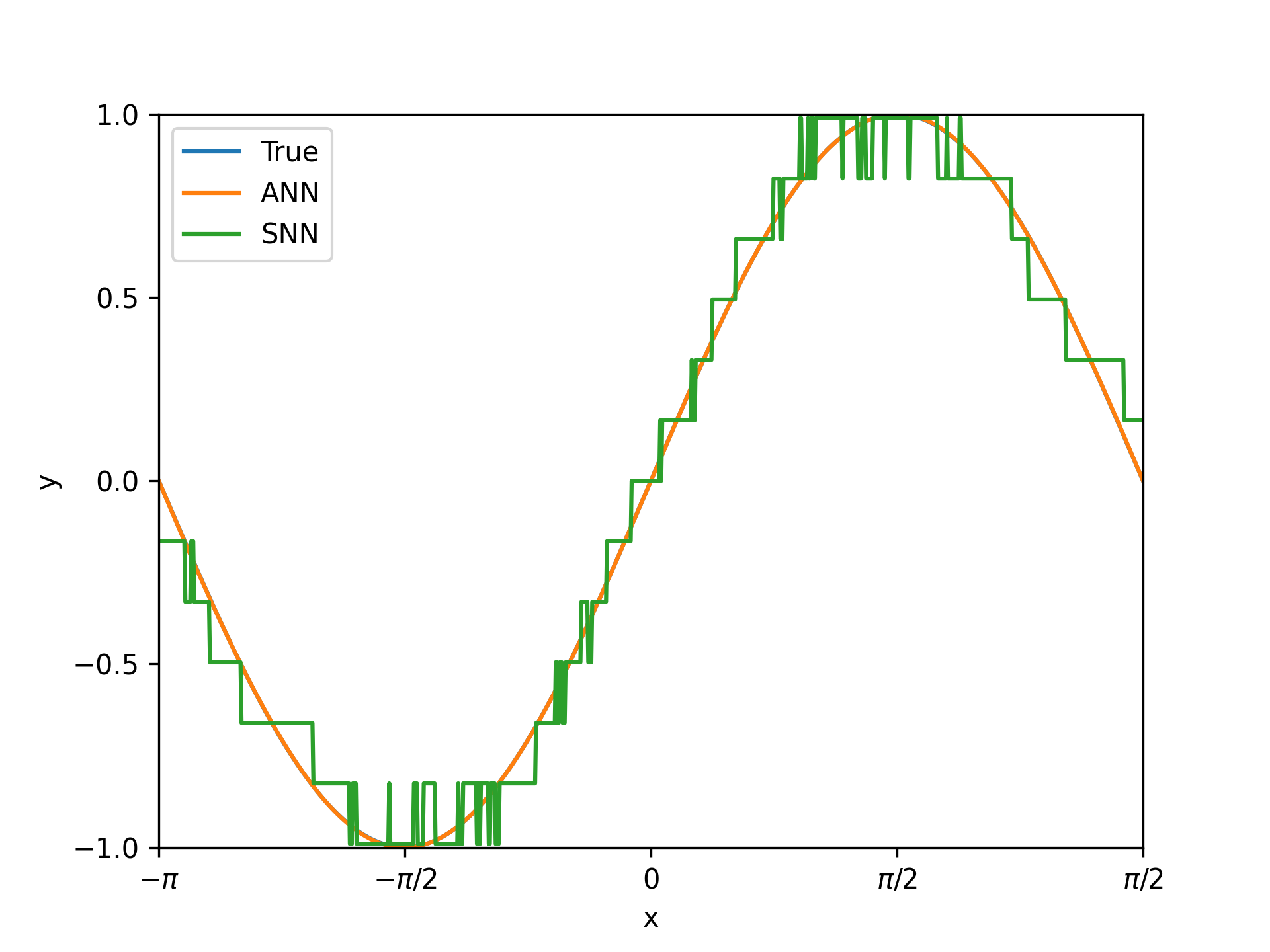}
        \caption{T=8}
    \end{subfigure}
    \begin{subfigure}{0.3\linewidth}
        \includegraphics[width=\linewidth]{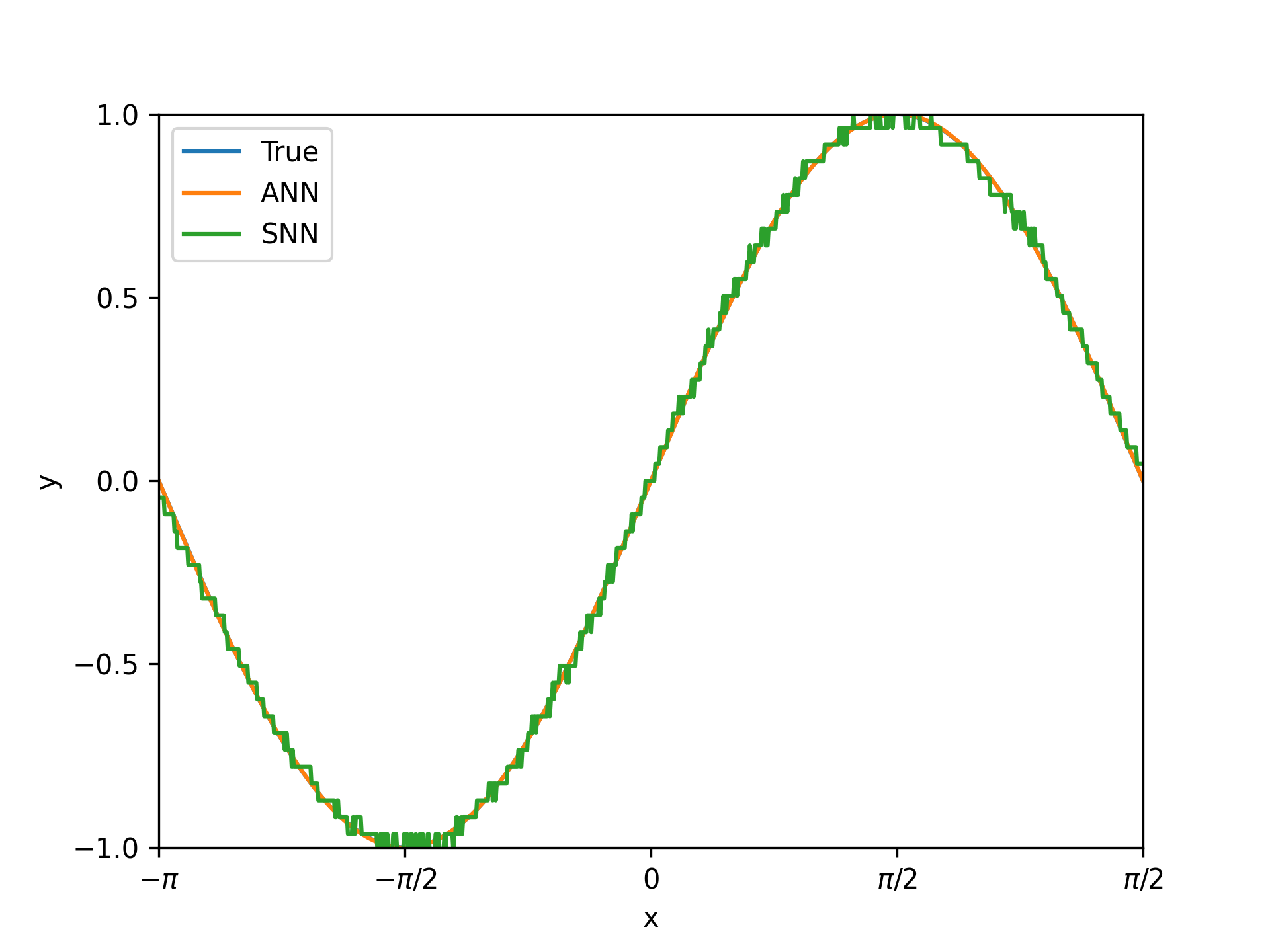}
        \caption{T=32}
    \end{subfigure}
    \begin{subfigure}{0.3\linewidth}
        \includegraphics[width=\linewidth]{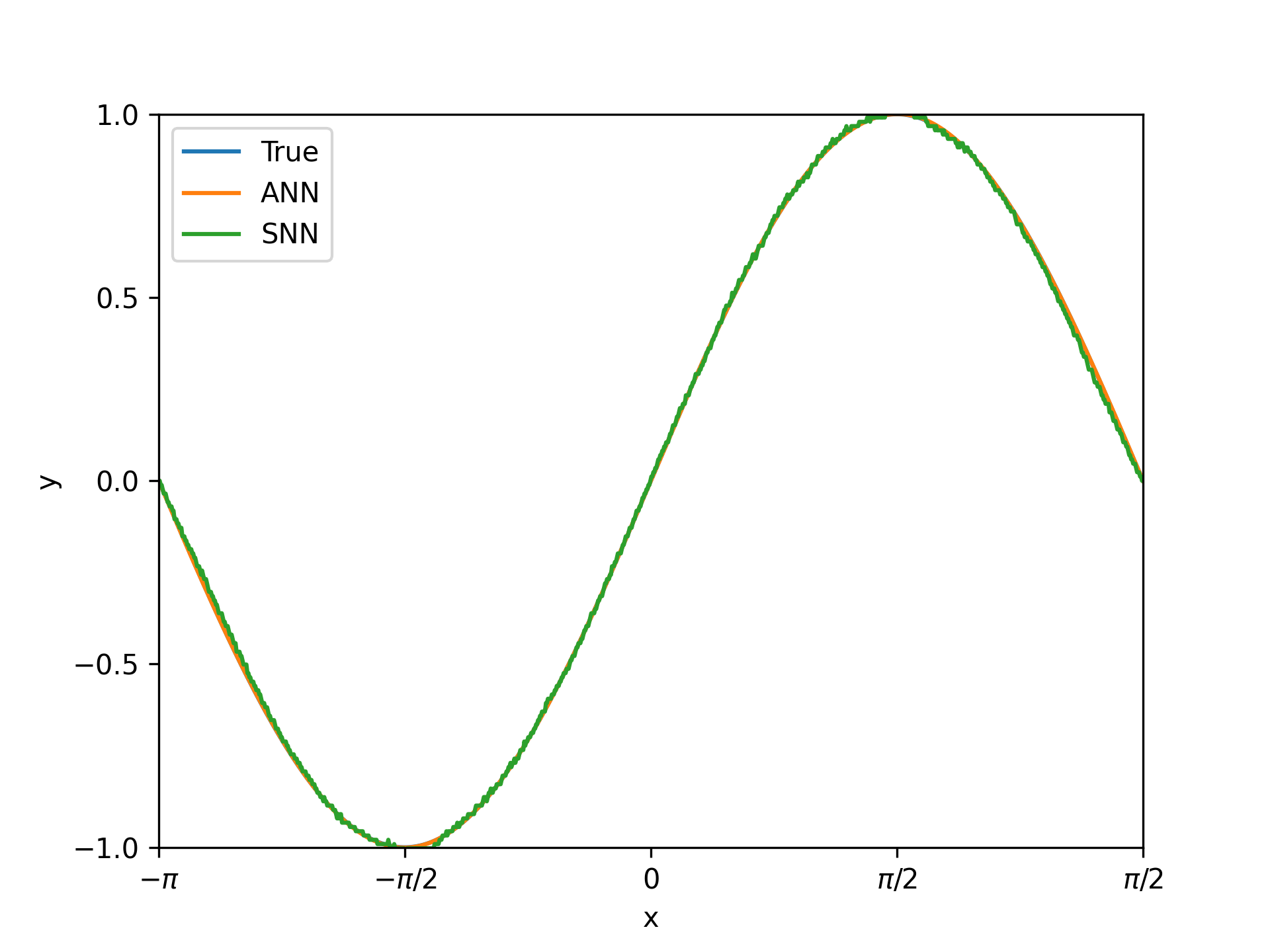}
        \caption{T=128}
    \end{subfigure}
    \caption{Results of converting an ANN, which is trained to approximate $\sin$, to SNN. The number of time steps is $T=8, 32, 128$ from left to right. The output of SNN is close to the ground truth, which is the output of ANN. With increasing T, the conversion error becomes smaller. When $T=128$, the SNN output curve is almost smooth and similar to the ground truth.}\label{fig:result:validation:sin-diff-T}
\end{figure}

\begin{figure}[htbp]
    \centering
    \begin{subfigure}{0.3\linewidth}
        \includegraphics[width=\linewidth]{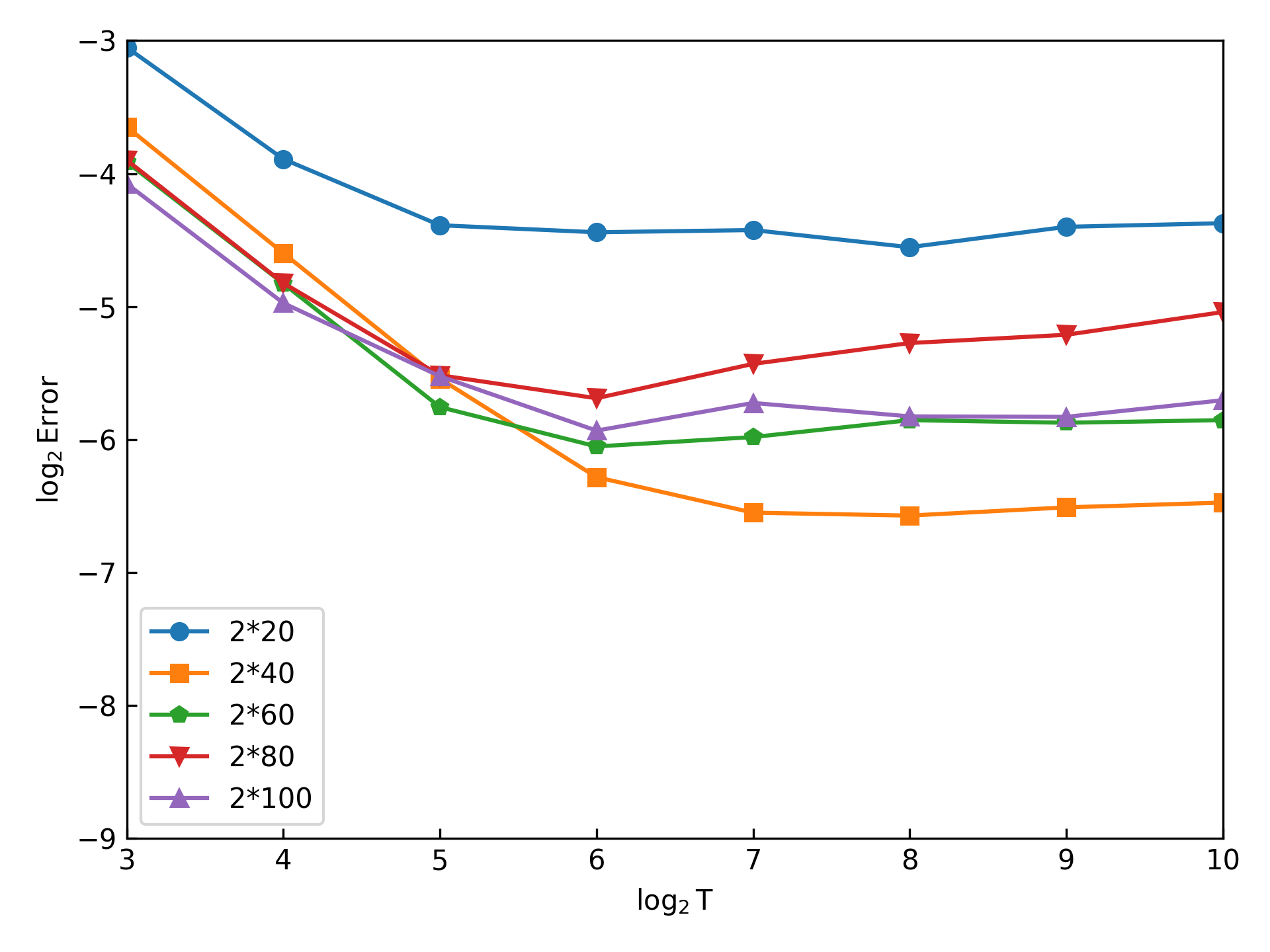}
        \caption{L=2}
    \end{subfigure}
    \begin{subfigure}{0.3\linewidth}
        \includegraphics[width=\linewidth]{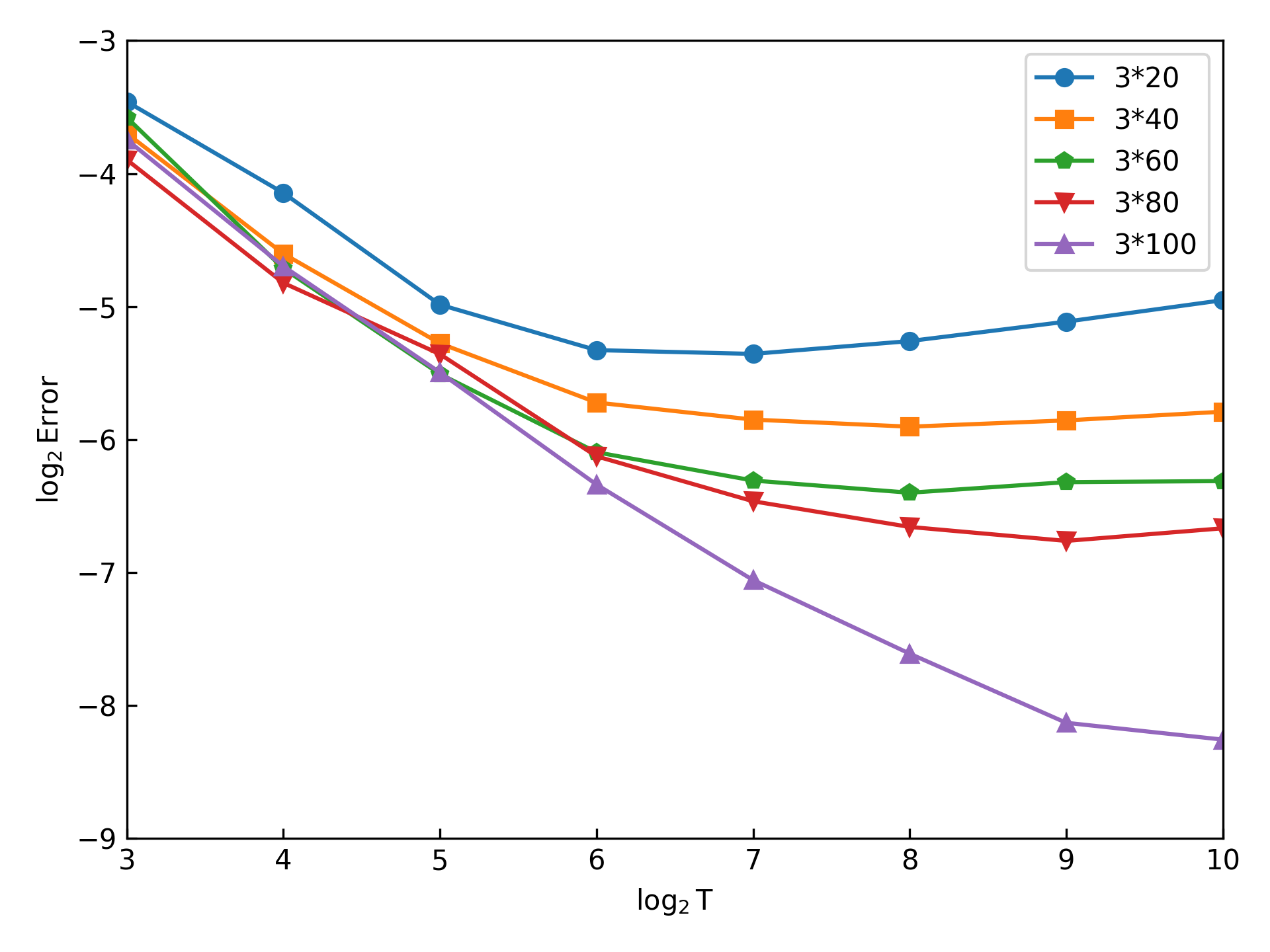}
        \caption{L=3}
    \end{subfigure}
    \begin{subfigure}{0.3\linewidth}
        \includegraphics[width=\linewidth]{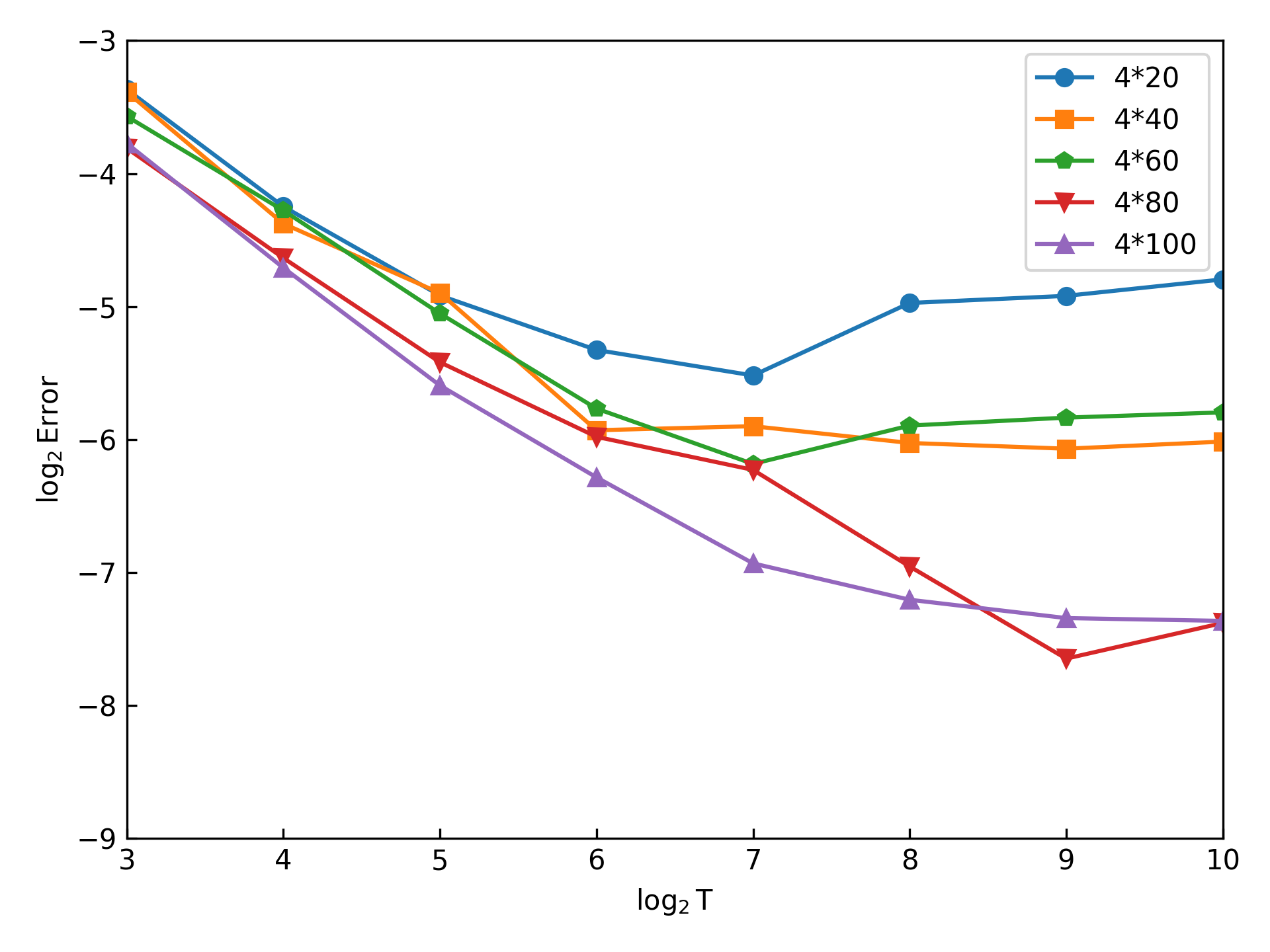}
        \caption{L=4}
    \end{subfigure}
    \caption{The number of layers are $L=2, 3, 4$ from left to right. The error here refers to $\e\at{n}$. When $T\leq64$, the conversion error follows $\e\at{n}\approx1/T$, and $\e\at{n}$ will get stable after $T$ is already very large.}\label{fig:result:validation:e-with-T}
\end{figure}

We first present an example of function regression: using neural networks to approximate the $\sin$ function. For the training dataset, the input is the uniform mesg points on $[-\pi.\pi]$, and the output is the values of $\sin$ on these points. The ANN model has two intermediate layers, each of them has 40 neurons. The activation function is $\tanh$ except for the last layer; there is no activation function for the last layer. The network is trained with the Adam optimizer until the training error is less then $1\time10^{-7}$. Then, we convert the ANN to SNN with advanced calibration with different numbers of time steps. The results are 
shown in Figure \ref{fig:result:validation:sin-diff-T}.
To further investigate the impact of $T$ on the conversion error, we train networks with different intermediate layer numbers $L=2,3,4$ and neuron numbers per layer $N = 20,40,60,80,100$. All the other setups are the same. We obtain the results shown in Figure \ref{fig:result:validation:e-with-T}.
We can find that the conversion error decreases with larger $T$ (more specifically, conversion error $\sim1/T$ where )  when $T<32$, but it becomes stable or larger after $T\geq32$. And for neural network with fixed depth, larger layer width usually leads to smaller conversion error. However, for fixed layer width, deeper neural networks will not bring significantly better conversion performance.

\begin{figure}[htbp]
    \centering
    \begin{subfigure}{0.3\linewidth}
        \includegraphics[width=\linewidth]{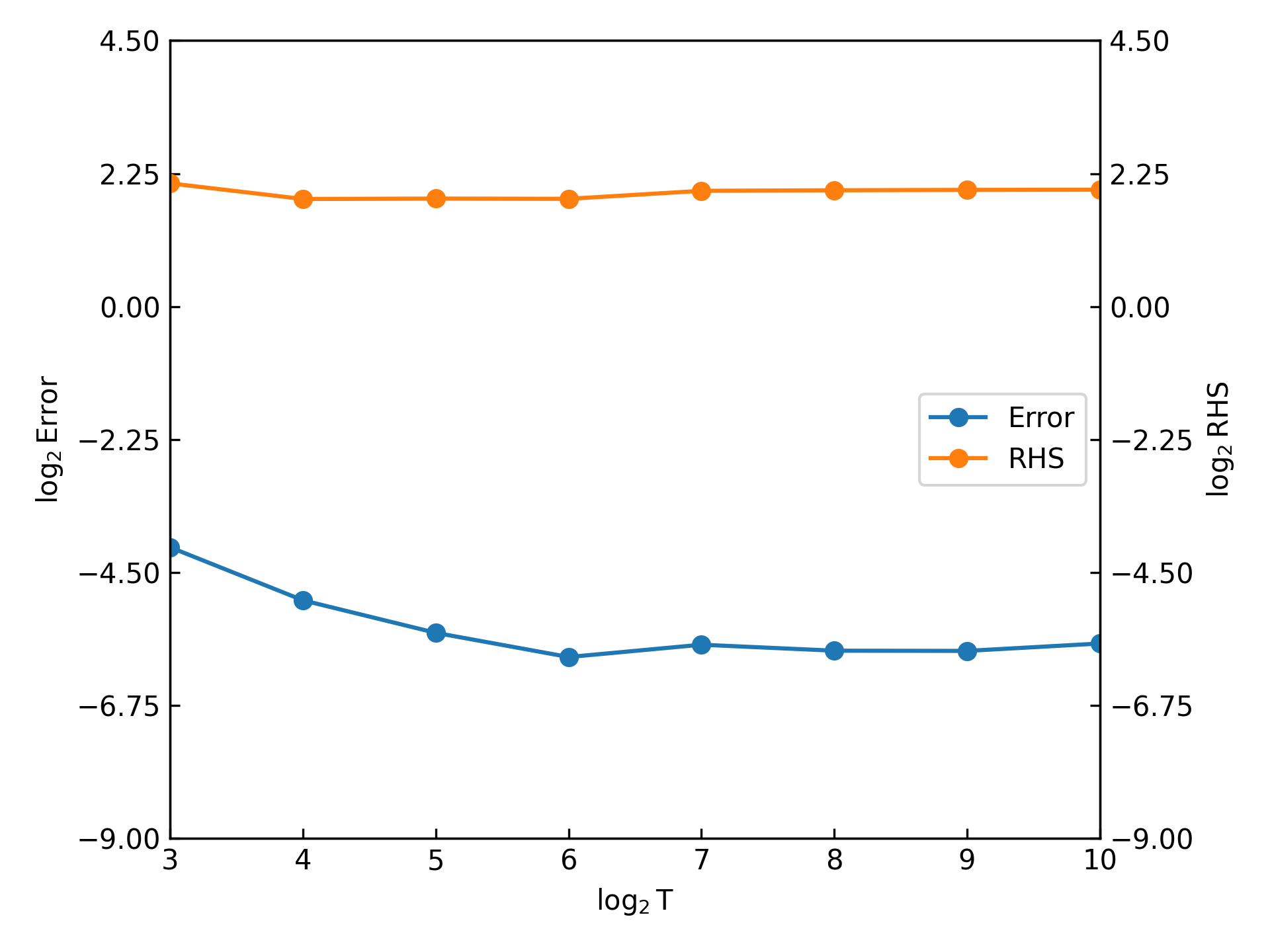}
        \caption{L=2}
    \end{subfigure}
    \begin{subfigure}{0.3\linewidth}
        \includegraphics[width=\linewidth]{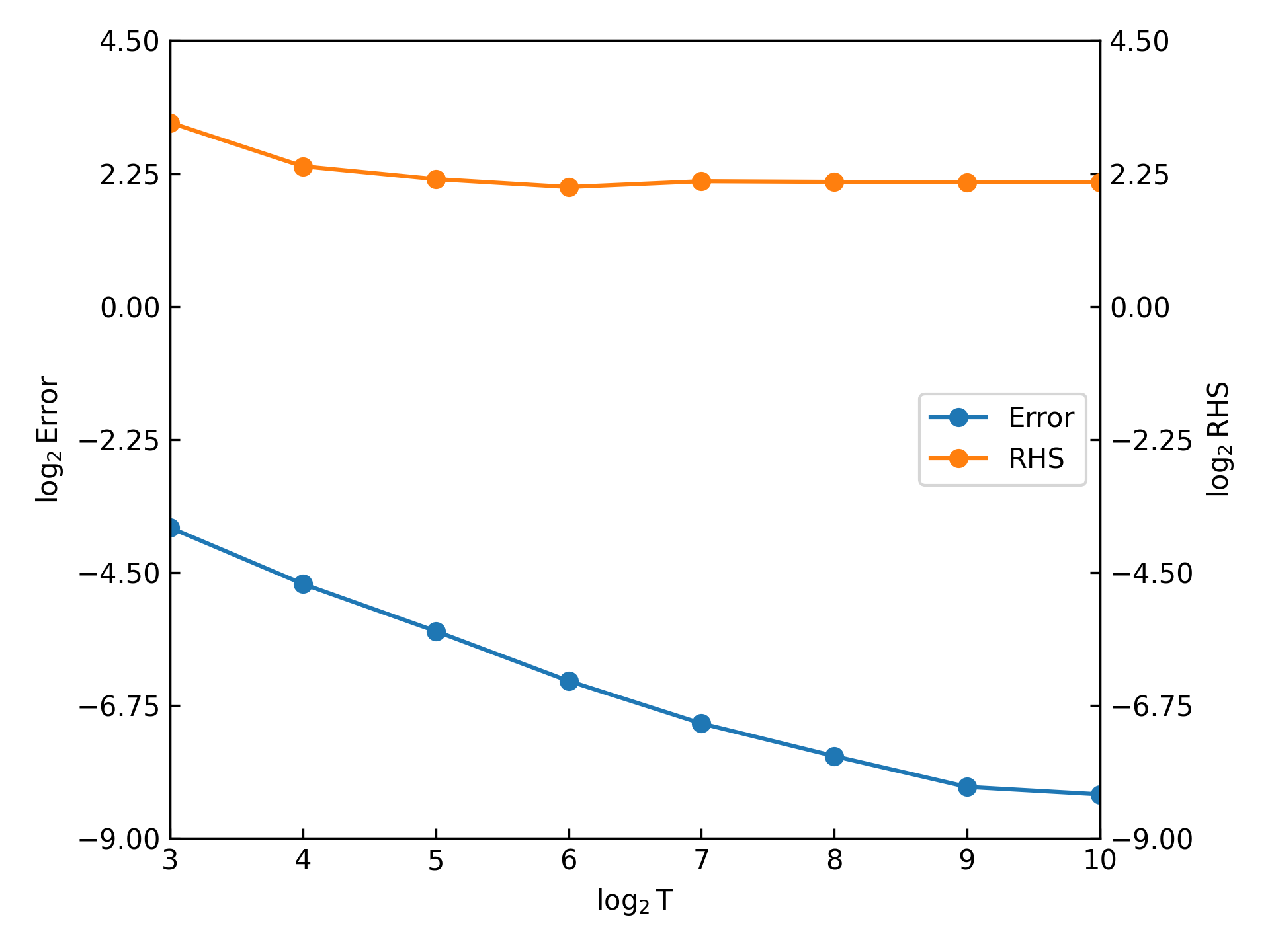}
        \caption{L=3}
    \end{subfigure}
    \begin{subfigure}{0.3\linewidth}
        \includegraphics[width=\linewidth]{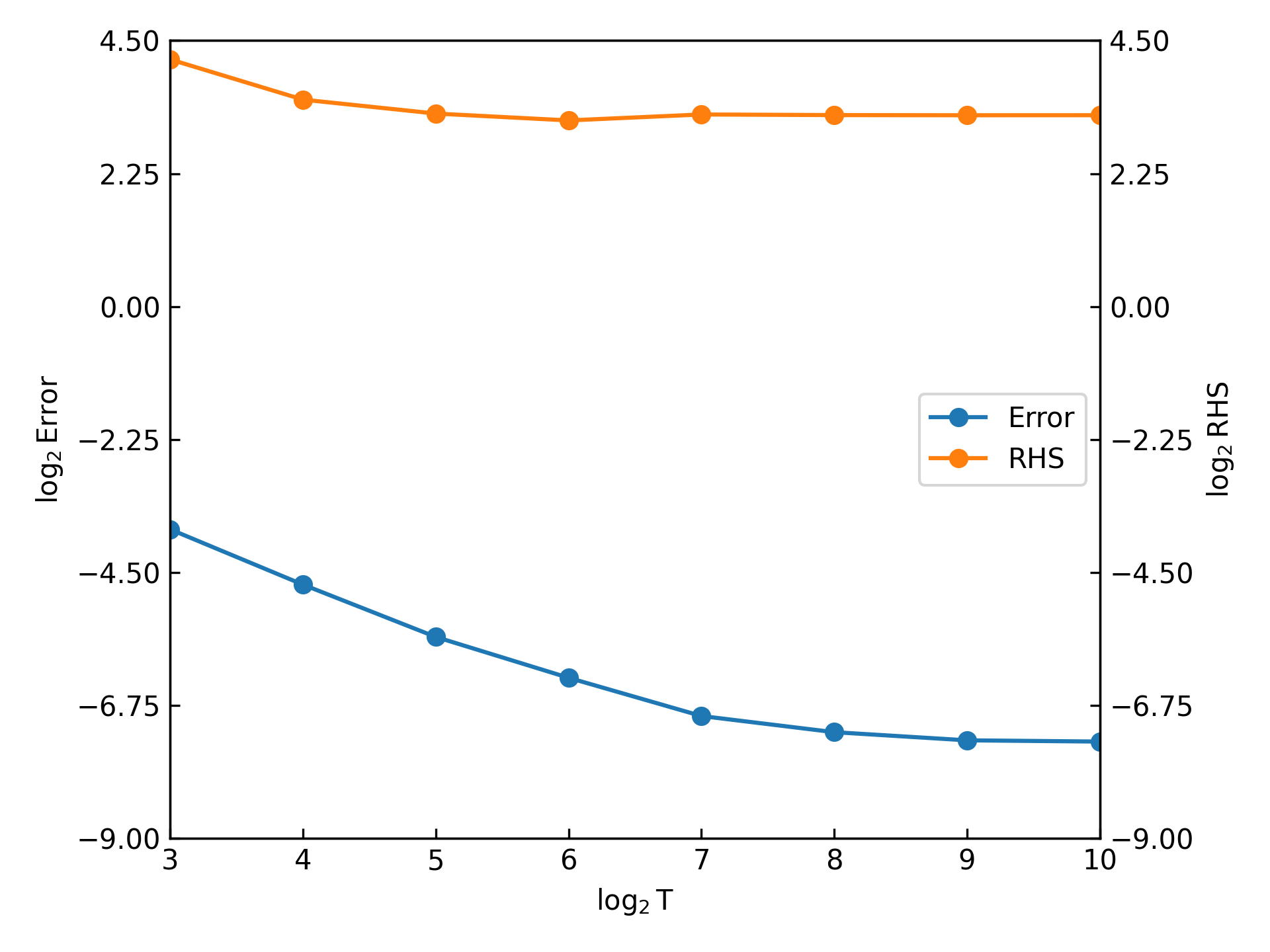}
        \caption{L=4}
    \end{subfigure}
    \caption{The number of layers are $L=2, 3, 4$ from left to right. The error here refers to $\e\at{n}$. RHS is defined as before. We can find that the total conversion error is smaller than the computed RHS and decrease with the RHS as well, which is an emprical validation of the Theorem \ref{thm:main}.}\label{fig:result:validation:e-with-ec}
\end{figure}

To validate the Theorem \ref{thm:main}, we need to compute $\e\att{l}\Hes\at{l}\e\at{l}$ for $l=1,2,\ldots,n$. Since $\Hes\at{l}$ are intractable, we can only use the identity matrix to replace it and obtain some qualitative results. That is, compute $\norm{\e\at{l}}^2$ instead of $\e\att{l}\Hes\at{l}\e\at{l}$. So the computed RHS is $\sum_{l=1}^n 2^{n-l+1}\e\att{l}\e\at{l}$. We train the ANN with layer numbers $L=2,3,4$ with $100$ neurons per layer. All the other setups are the same. The results are shown in Figure \ref{fig:result:validation:e-with-ec}.
Although the RHS term is not exactly the same as in \ref{thm:main}, the trend agrees with our statement, that conversion error decreases as the RHS term does.

\subsection{PINNs}
To show the power of SNN conversion in regression tasks, we train PINNs, a MLP-based neural network which can solve PDEs, and convert them to SNNs. 
We present results for the Poisson equation, the diffusion-reaction equation, the wave equation, the Burgers equation, and the Navier-Stokes equations.

\begin{figure}[htbp]
    \centering
    \begin{subfigure}{0.245\linewidth}
        \includegraphics[width=\linewidth]{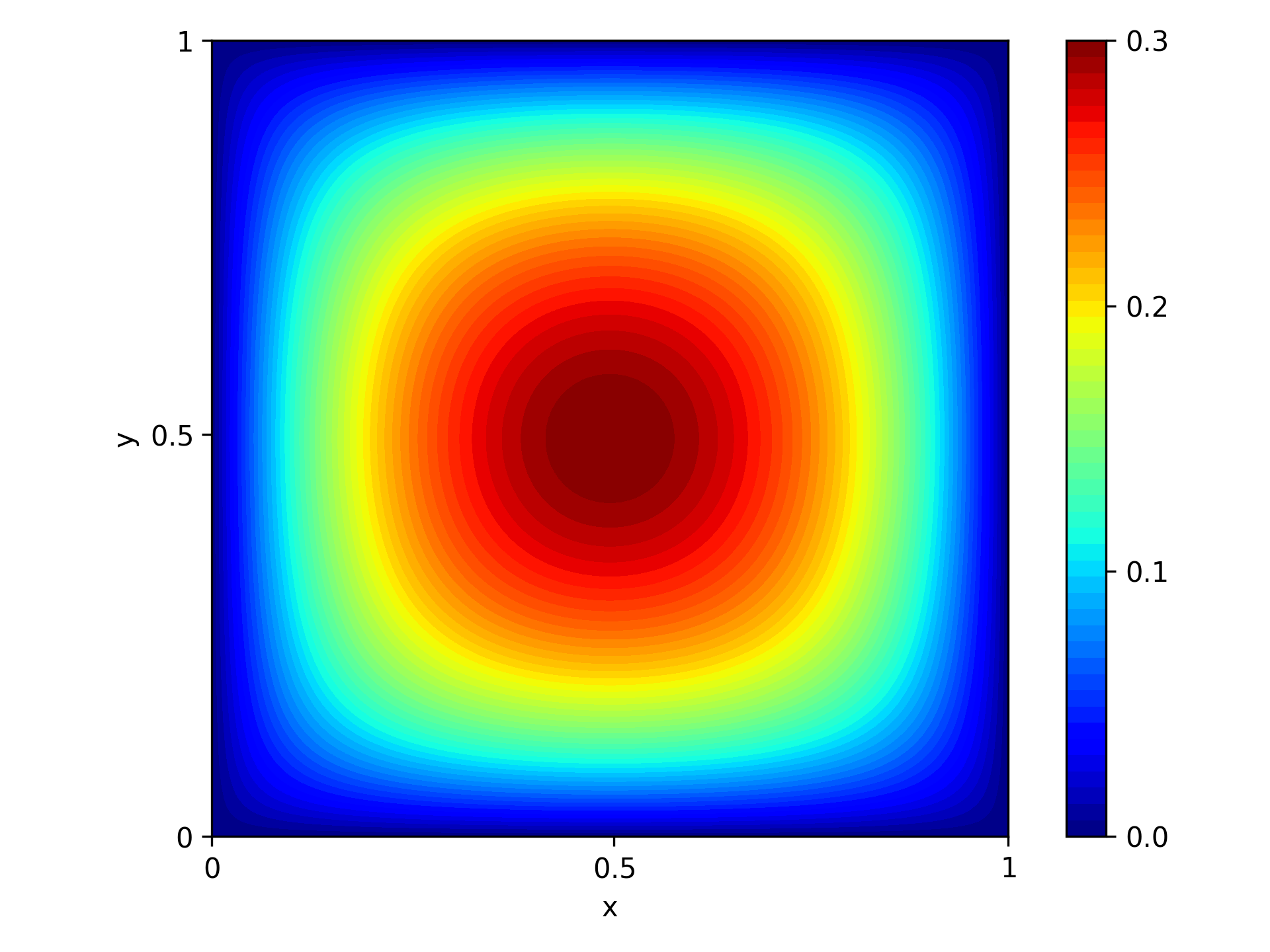}
        \caption{Reference solution}\label{fig:result:pinn-poisson:true}
    \end{subfigure}
    \begin{subfigure}{0.245\linewidth}
        \includegraphics[width=\linewidth]{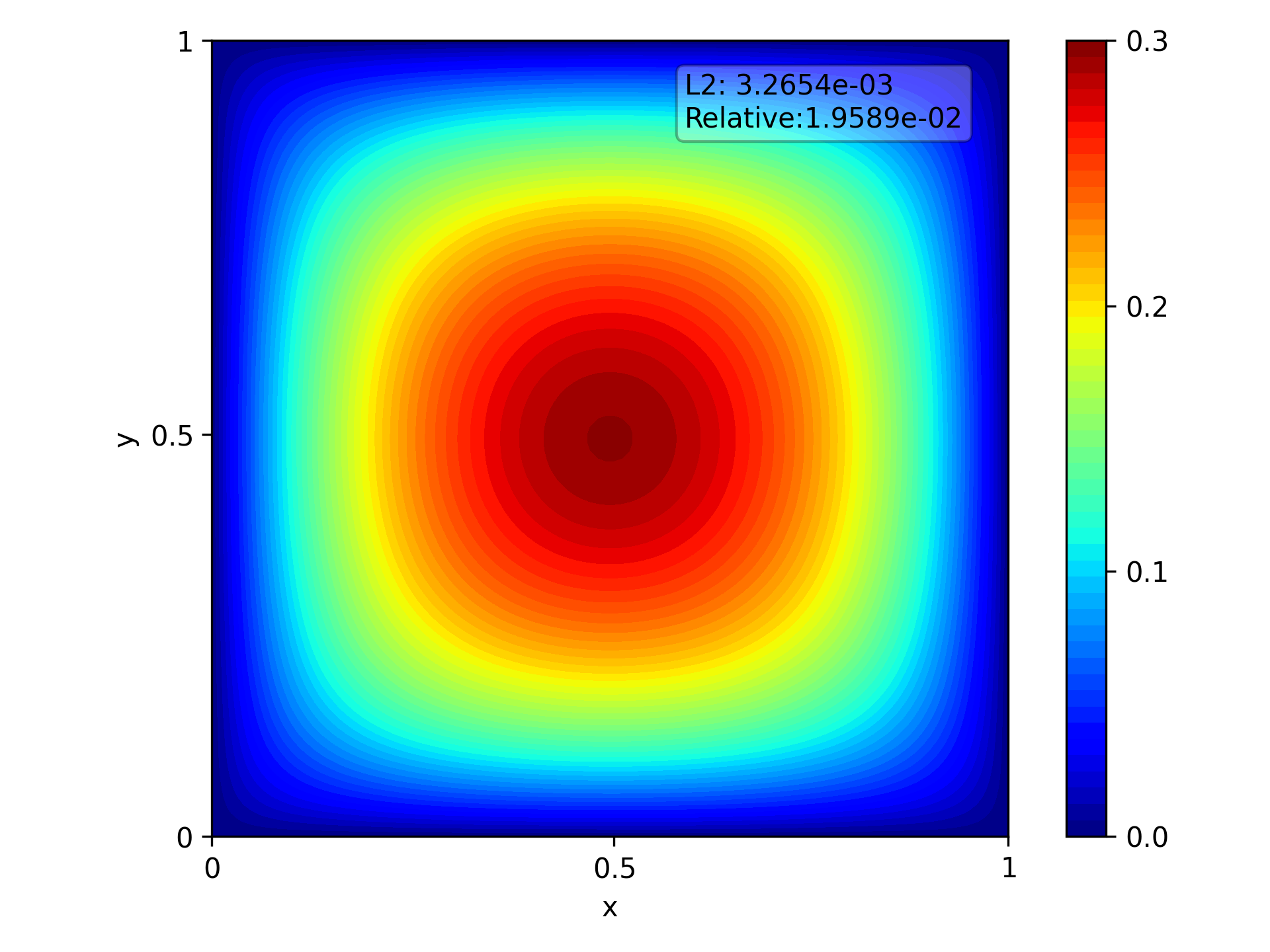}
        \caption{PINN result}\label{fig:result:pinn-poisson:ann}
    \end{subfigure}
    \begin{subfigure}{0.245\linewidth}
        \includegraphics[width=\linewidth]{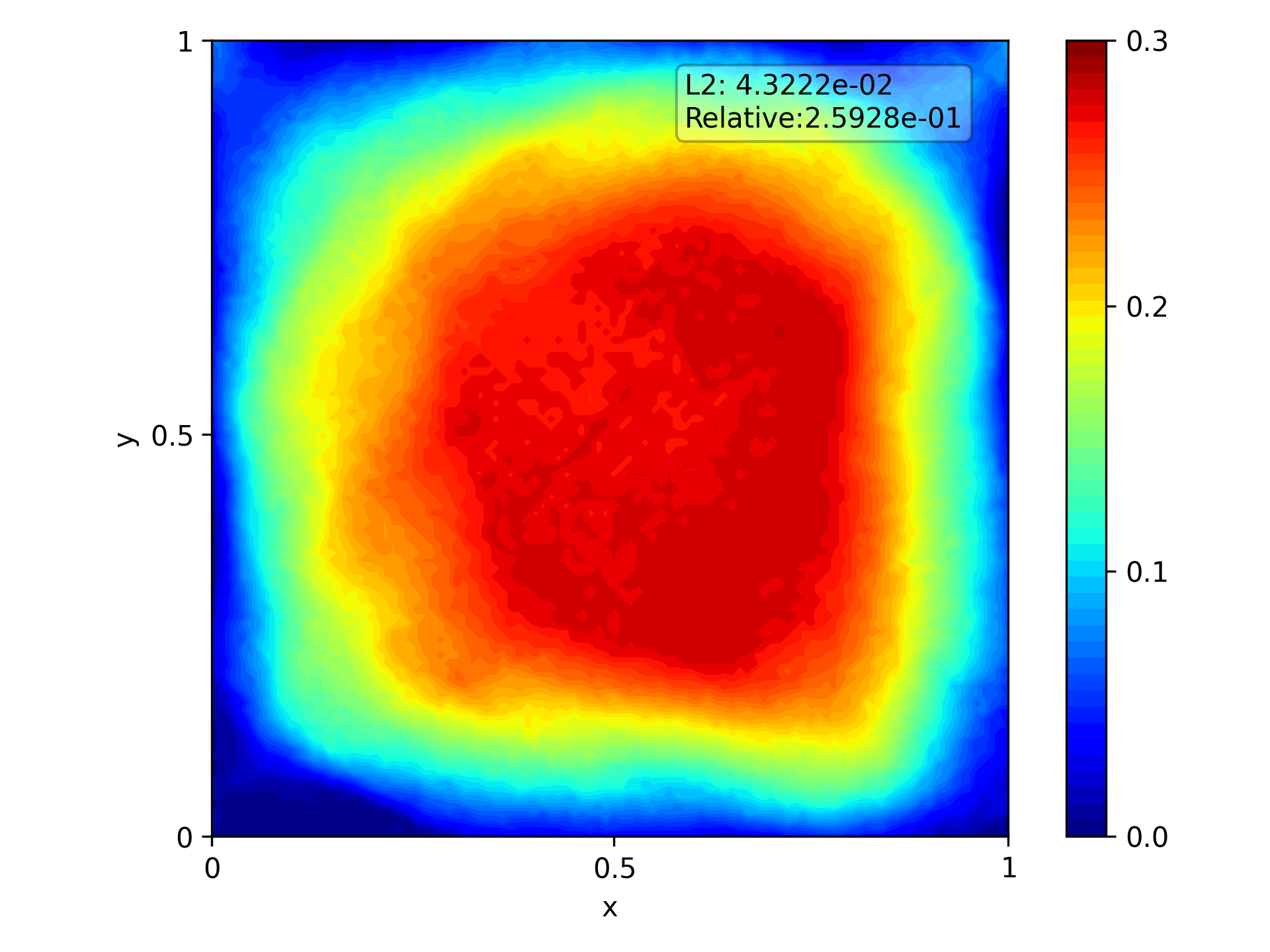}
        \caption{Conversion w/o calibration}\label{fig:result:pinn-poisson:snn_none}
    \end{subfigure}
    \begin{subfigure}{0.245\linewidth}
        \includegraphics[width=\linewidth]{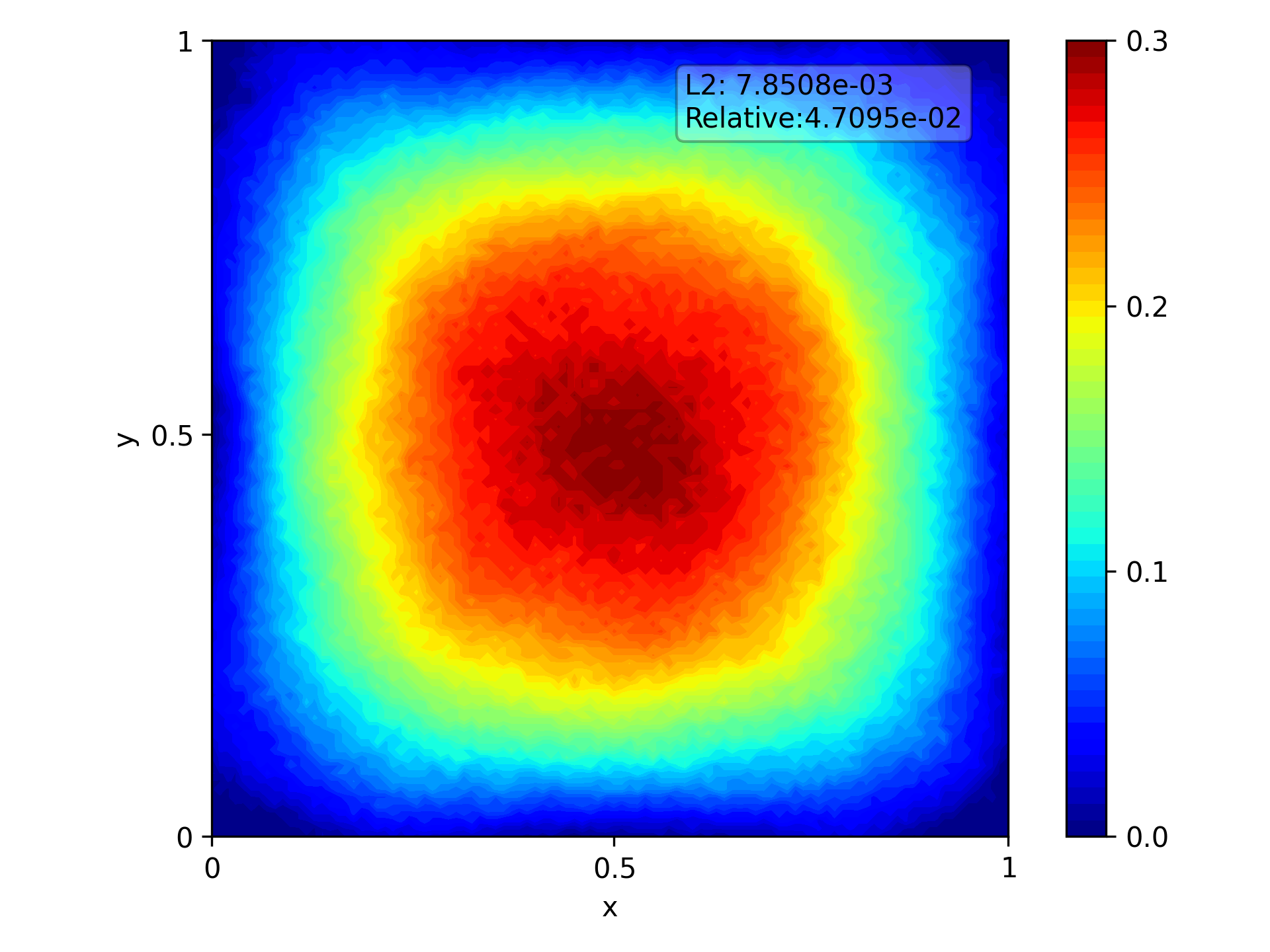}
        \caption{Conversion w/ calibration}\label{fig:result:pinn-poisson:snn_advanced}
    \end{subfigure}
    \caption{Poisson equation: The results of converting a PINN solving the poisson equation (\ref{eqn:result:poisson}). Figure \ref{fig:result:pinn-poisson:true} is the reference solution. Figure \ref{fig:result:pinn-poisson:ann} is the PINN result. Figure \ref{fig:result:pinn-poisson:snn_none} is the result of the SNN converted from the PINN without using calibration. Figure \ref{fig:result:pinn-poisson:snn_none} is the result of the SNN converted from the PINN using calibration. Here the L2 error and relative error are defined as $\norm{\x\at{n}-\bar{\s}\at{n}}_2$ and $\norm{\x\at{n}-\bar{\s}\at{n}}_2/\norm{\x\at{n}}_2$, where $\x\at{n}$ is the reference solution and $\bar{\s}\at{n}$ is the neural network output. $\norm{\cdot}_2$ is the $l^2$ norm, which is the root of mean square error.}\label{fig:result:pinn-poisson}
\end{figure}

\paragraph{Poisson equation} The Poisson equation is often used to describe the potential field in electromagnetic theory. Here we solve the following boundary value problem of Poisson equation
\begin{equation}
\begin{aligned}
    -\Delta u(x) &= 1,\quad x\in\Omega=[-1,1]\times[-1,1]\\
    u(x) &= 0,\quad x\in\partial\Omega
\end{aligned}\label{eqn:result:poisson}
\end{equation}
with PINN. The network has $3$ intermediate layers, each of which has 100 neurons. The activation function is $\tanh$ except for the last layer. The network is trained for 50,000 epochs. Then we convert it into SNN.  The results are shown in Figure \ref{fig:result:pinn-poisson}. 
We observe that the SNN converted with calibration can achieve the same magnitude of error as the PINN evaluated as an ANN, while the SNN converted without using calibration can just obtain a rough shape of the solution and has much larger error.

\begin{figure}[htbp]
    \centering
    \begin{subfigure}{0.245\linewidth}
        \includegraphics[width=\linewidth]{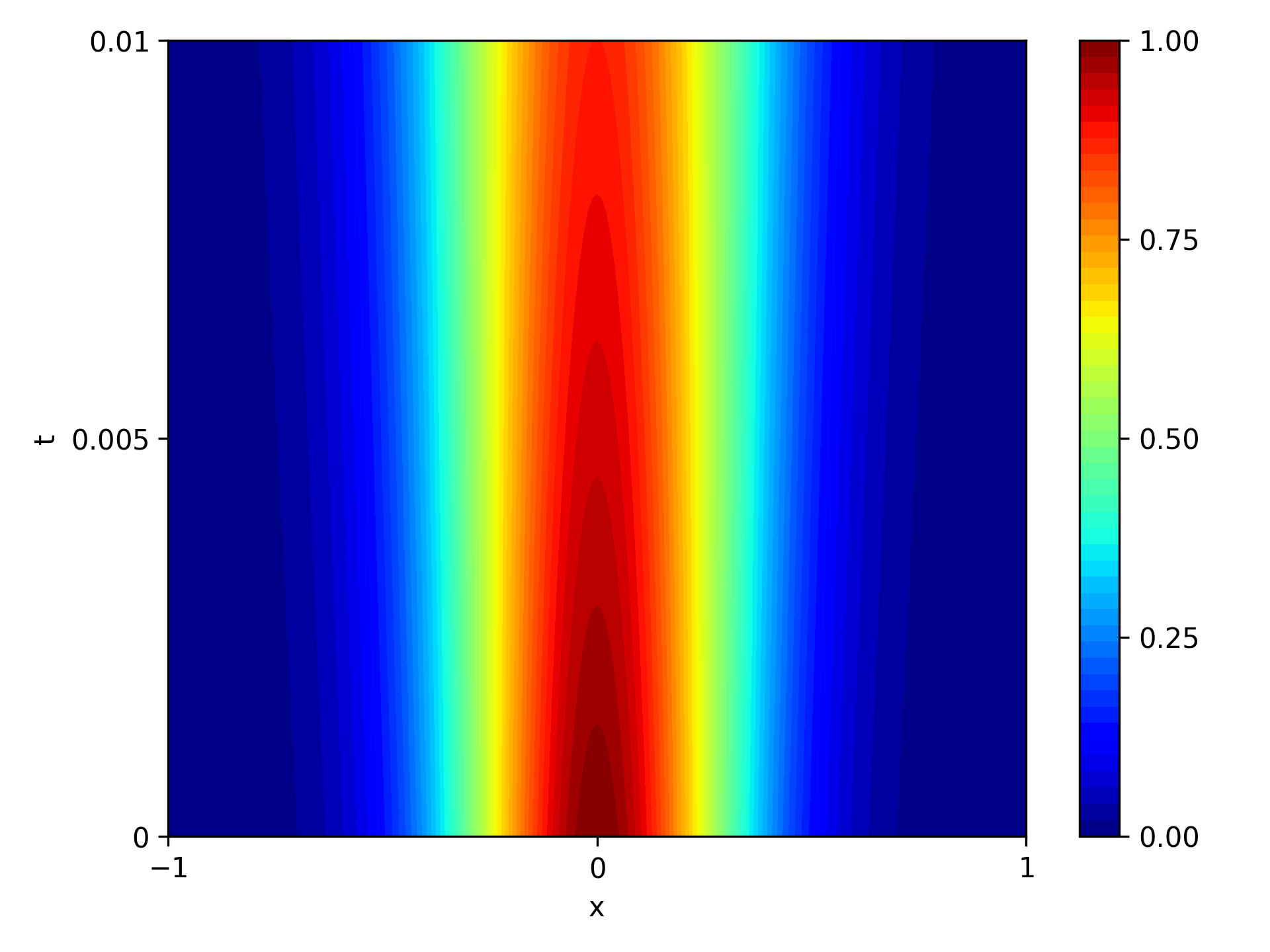}
        \caption{Reference solution}\label{fig:result:pinn-heat_nonlinear:true}
    \end{subfigure}
    \begin{subfigure}{0.245\linewidth}
        \includegraphics[width=\linewidth]{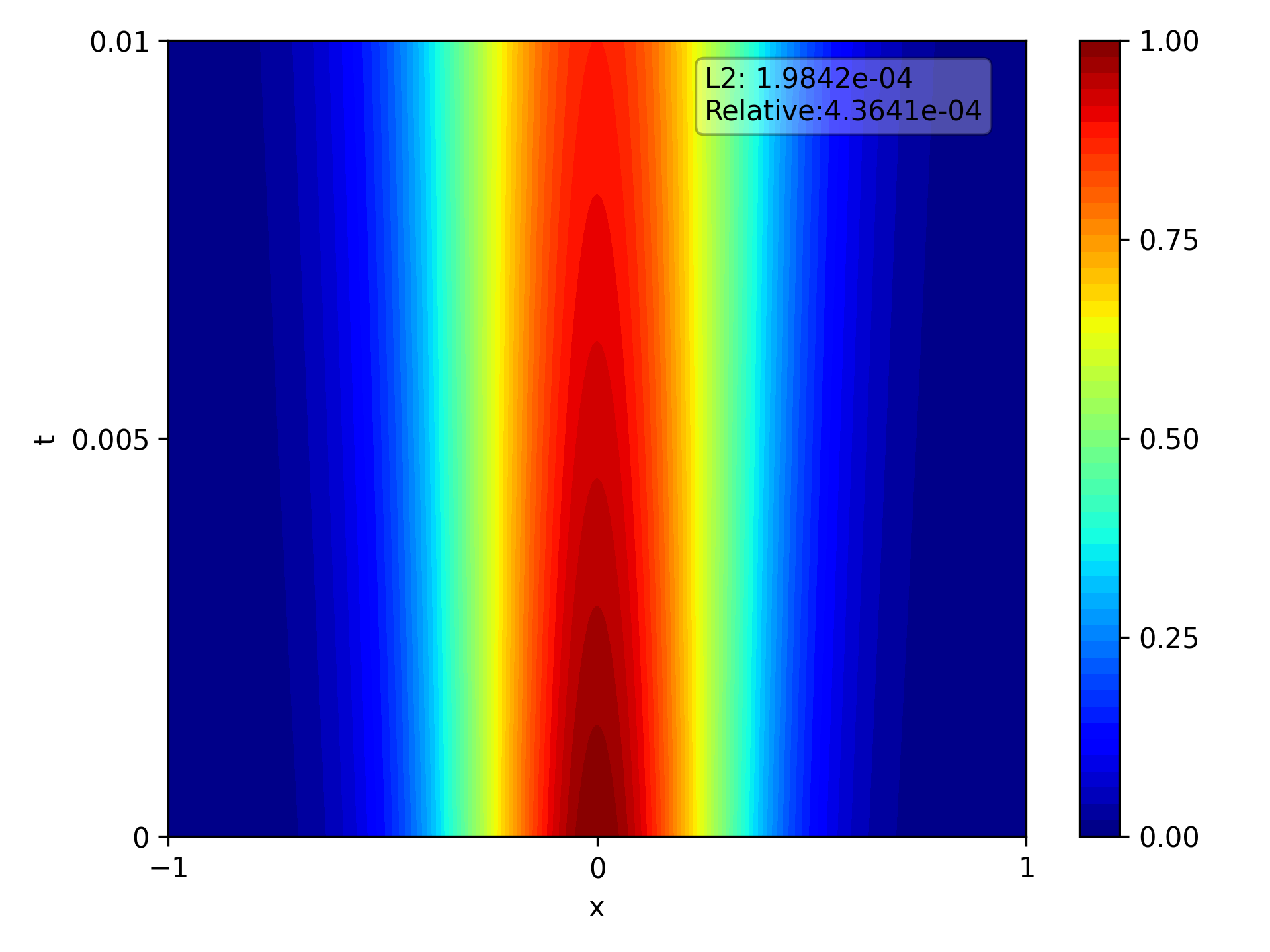}
        \caption{PINN result}\label{fig:result:pinn-heat_nonlinear:ann}
    \end{subfigure}
    \begin{subfigure}{0.245\linewidth}
        \includegraphics[width=\linewidth]{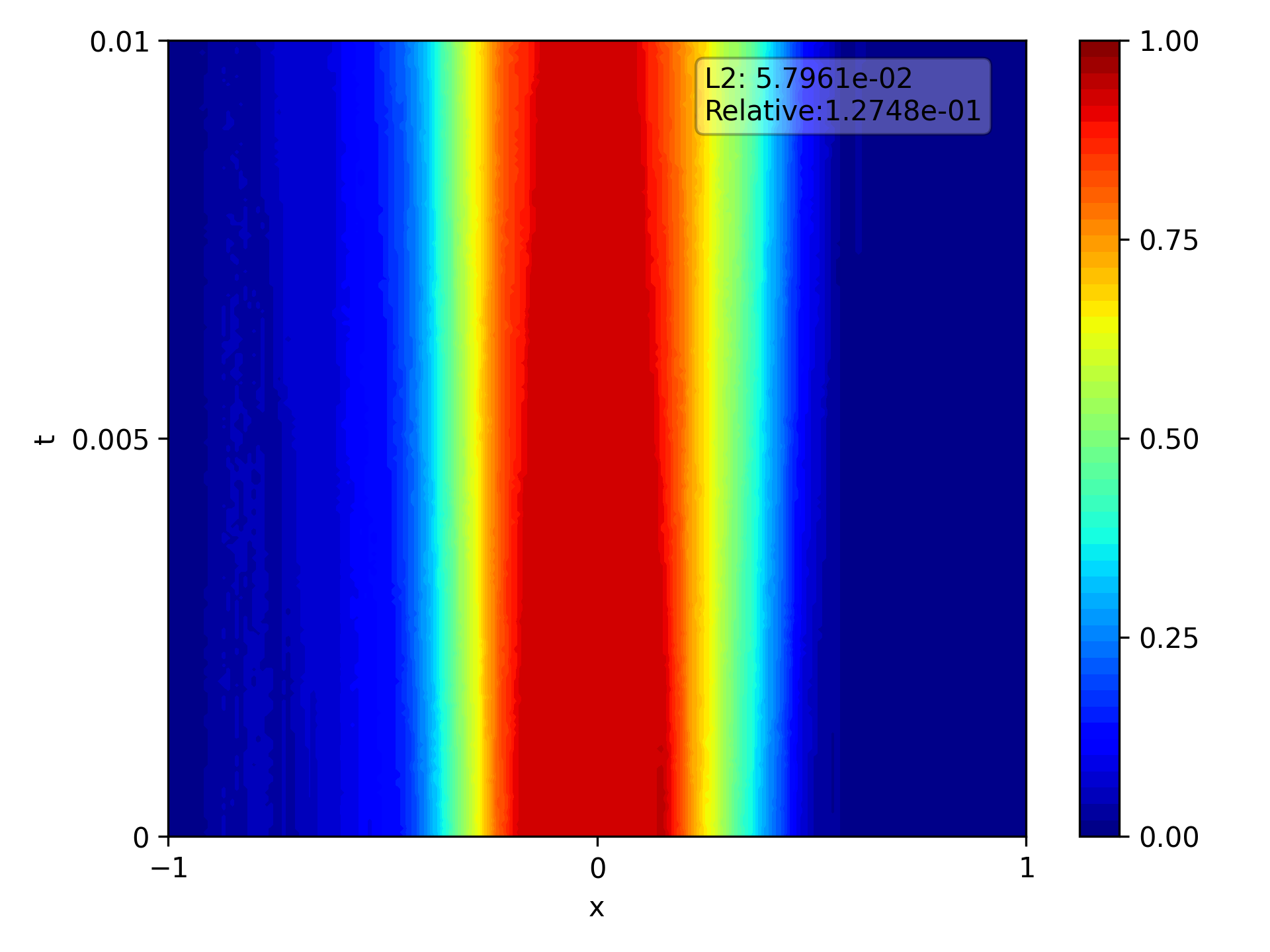}
        \caption{Conversion w/o calibration}\label{fig:result:pinn-heat_nonlinear:snn_none}
    \end{subfigure}
    \begin{subfigure}{0.245\linewidth}
        \includegraphics[width=\linewidth]{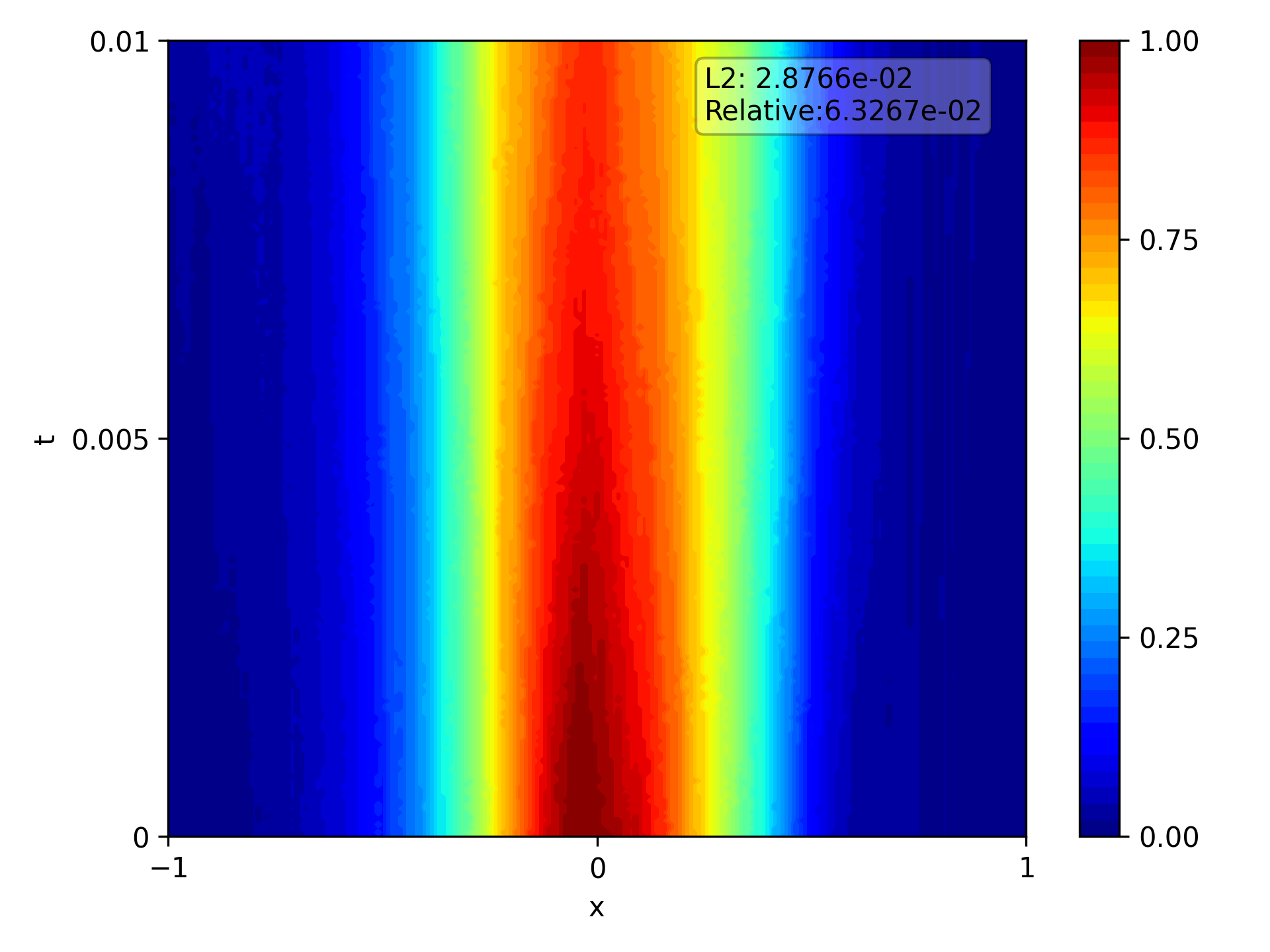}
        \caption{Conversion w/ calibration}\label{fig:result:pinn-heat_nonlinear:snn_advanced}
    \end{subfigure}
    \caption{Reaction-diffusion equation: The results of converting a PINN solving the nonlinear heat equation (\ref{eqn:result:heat_nonlinear}). Figure \ref{fig:result:pinn-heat_nonlinear:true} is the reference solution. Figure \ref{fig:result:pinn-heat_nonlinear:ann} is the PINN result. Figure \ref{fig:result:pinn-heat_nonlinear:snn_none} is the result of the SNN converted from the PINN without using calibration. Figure \ref{fig:result:pinn-heat_nonlinear:snn_none} is the result of the SNN converted from the PINN using calibration.}\label{fig:result:pinn-heat_nonlinear}
\end{figure}

\paragraph{Diffusion-reaction equation} This equation models reactive transport in physical and biologicak systems. Here we use PINN to solve the diffusion-reaction equation with the following initial condition:
\begin{equation}
\begin{aligned}
    u_t-u_{xx} &= ku^2,\quad x\in\Omega=[-1,1]\\
    u(x, 0) &= \exp(-\frac{x^2}{2\sigma^2})
\end{aligned}\label{eqn:result:heat_nonlinear}
\end{equation}
where $k=1$, $\sigma=0.25$ up to time $T=0.01$. The network has $3$ intermediate layers, each of which has 100 neurons. The activation function is $\tanh$ except for the last layer. The network is trained for 100,000 epochs. Then we convert it into SNN. The results are shown 
in \ref{fig:result:pinn-heat_nonlinear}.

\begin{figure}[htbp]
    \centering
    \begin{subfigure}{0.245\linewidth}
        \includegraphics[width=\linewidth]{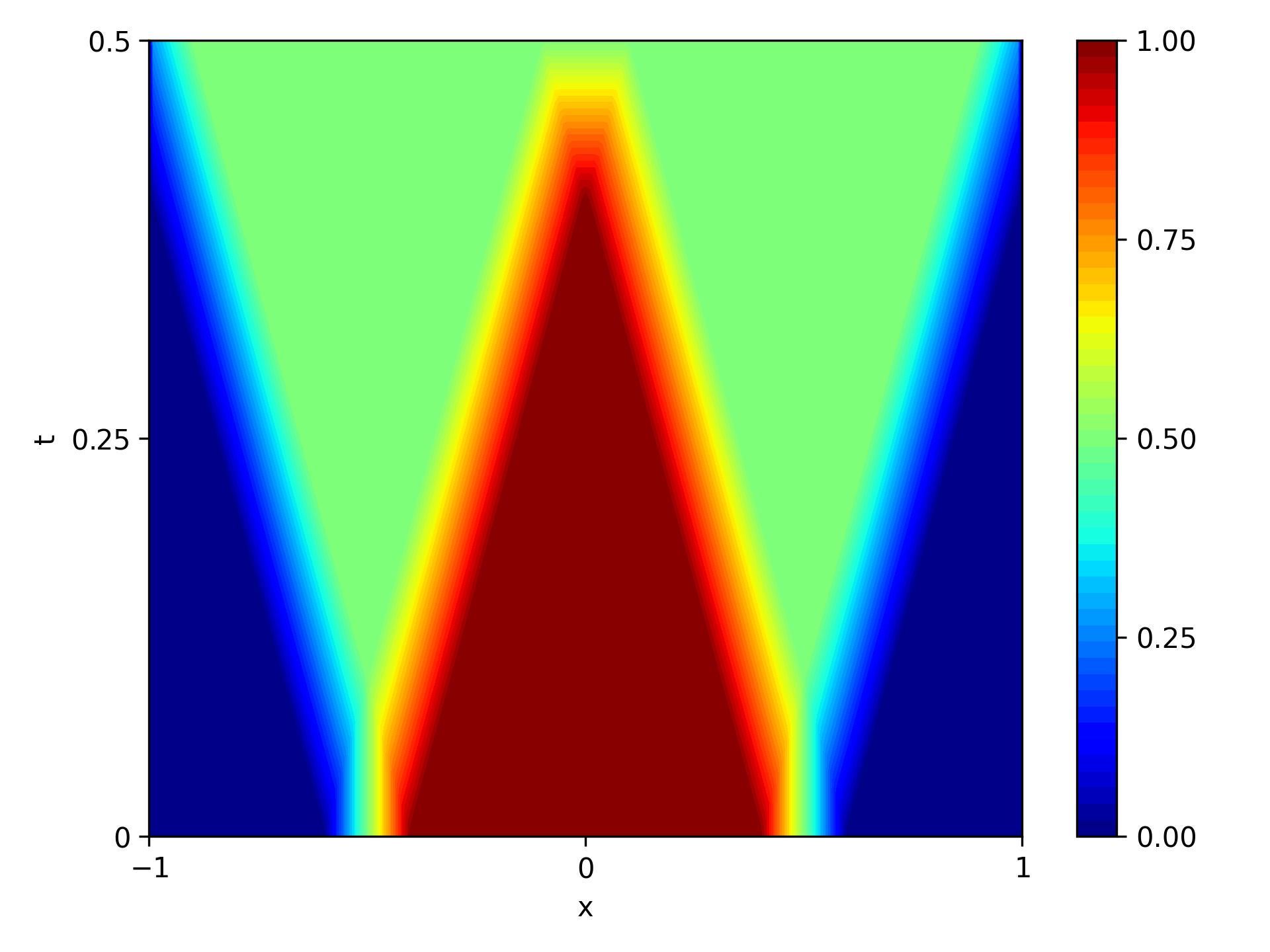}
        \caption{Reference solution}\label{fig:result:pinn-wave:true}
    \end{subfigure}
    \begin{subfigure}{0.245\linewidth}
        \includegraphics[width=\linewidth]{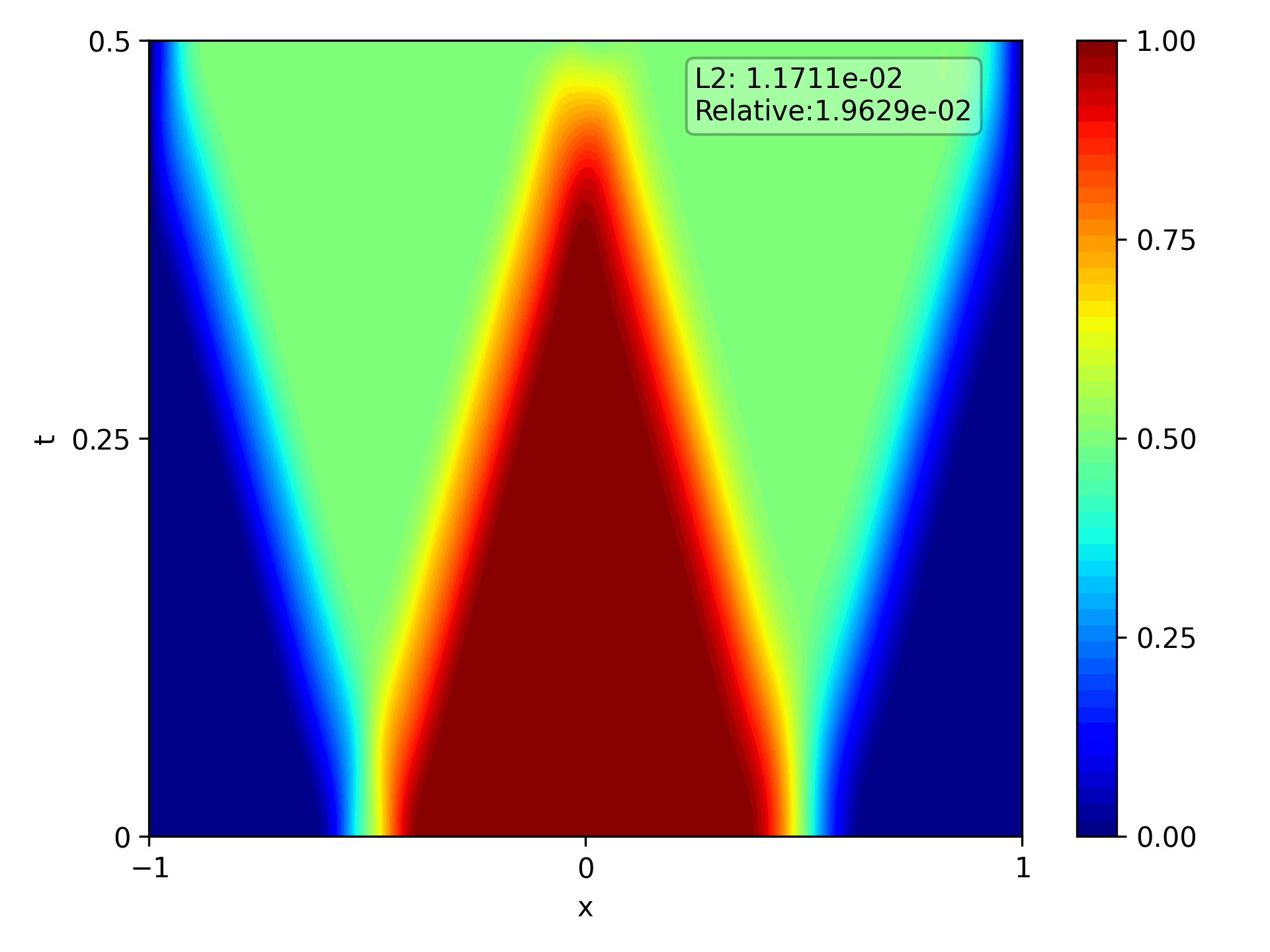}
        \caption{PINN result}\label{fig:result:pinn-wave:ann}
    \end{subfigure}
    \begin{subfigure}{0.245\linewidth}
        \includegraphics[width=\linewidth]{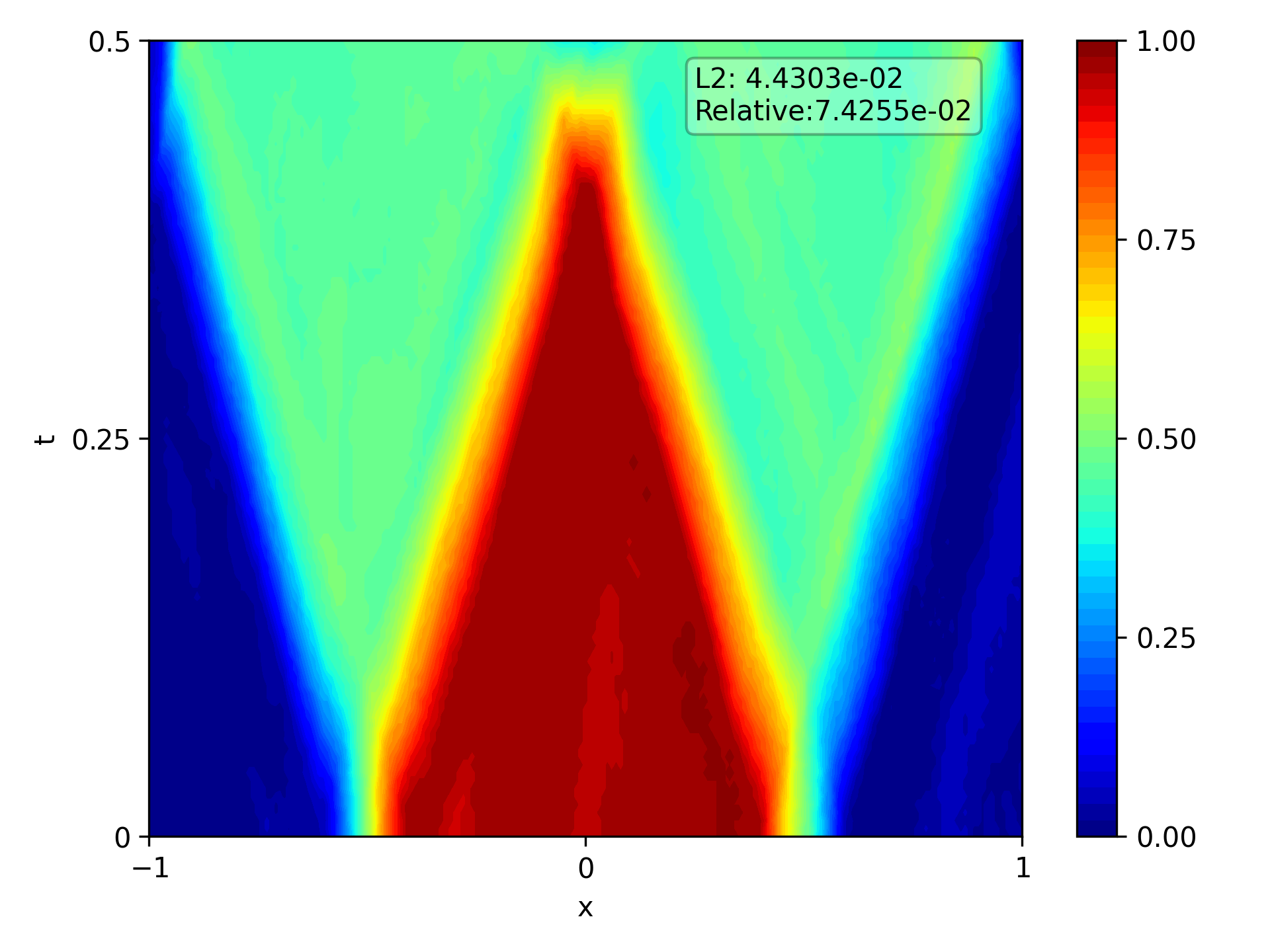}
        \caption{Conversion w/o calibration}\label{fig:result:pinn-wave:snn_none}
    \end{subfigure}
    \begin{subfigure}{0.245\linewidth}
        \includegraphics[width=\linewidth]{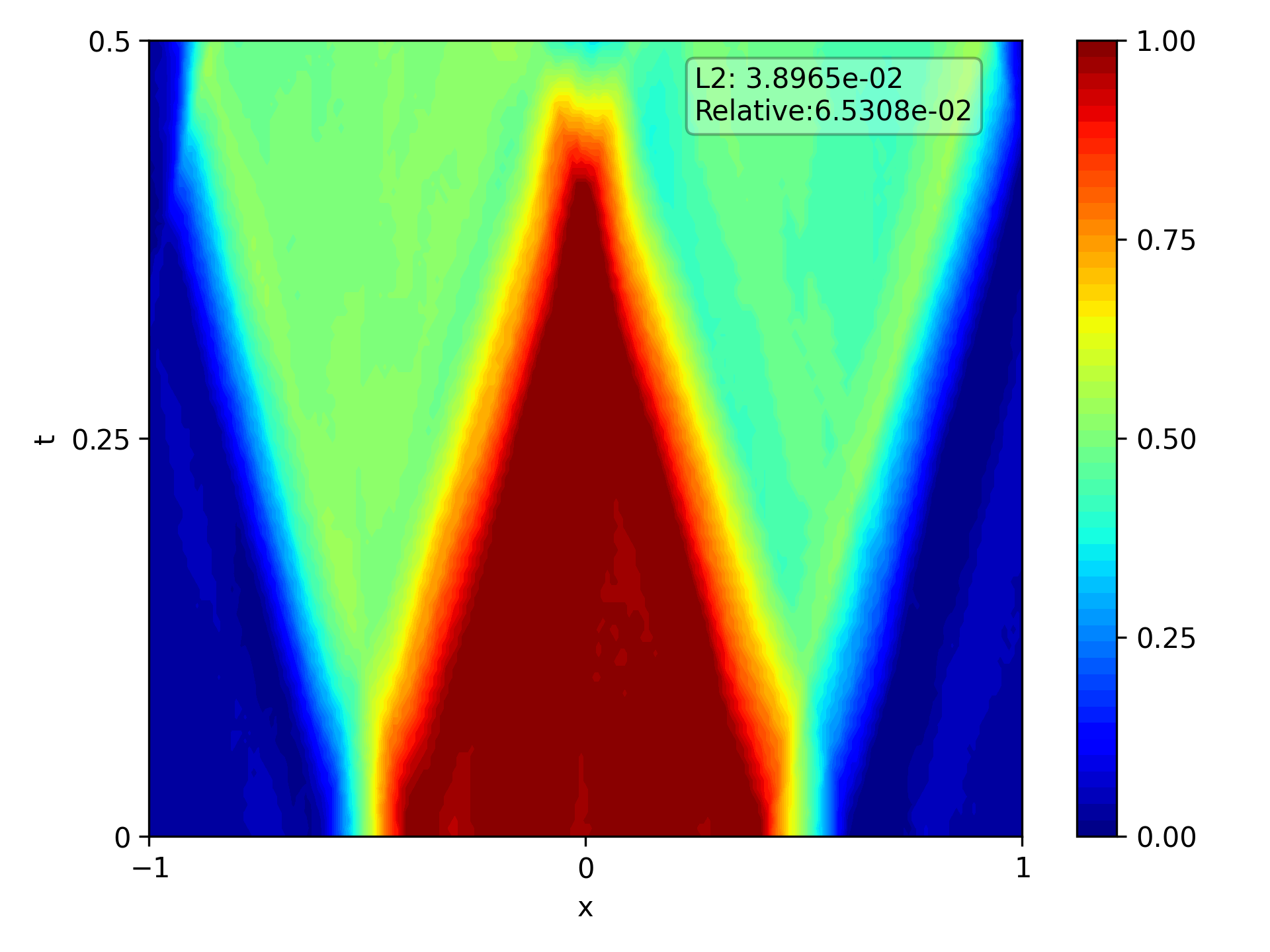}
        \caption{Conversion w/ calibration}\label{fig:result:pinn-wave:snn_advanced}
    \end{subfigure}
    \caption{Wave equation: The results of converting a PINN solving the wave equation (\ref{eqn:result:wave}). Figure \ref{fig:result:pinn-wave:true} is the reference solution. Figure \ref{fig:result:pinn-wave:ann} is the PINN result. Figure \ref{fig:result:pinn-wave:snn_none} is the result of the SNN converted from the PINN without using calibration. Figure \ref{fig:result:pinn-wave:snn_none} is the result of the SNN converted from the PINN using calibration.}\label{fig:result:pinn-wave}
\end{figure}

\paragraph{Wave equation}  Here we use PINN to solve the a wave equation with the following initial and boundary conditions:
\begin{equation}
\begin{aligned}
    u_{tt}-u_{xx} &= 0,\quad x\in\Omega=[-1,1]\\
    u(x, 0) &= \left\{\begin{aligned}
         & 1\quad x\in[-0.245, 0.245]\\
         & 0\quad x\in[-1, -0.6]\cap[0.6,1]\\
         & \mathrm{linear}\quad \mathrm{otherwise}
    \end{aligned}\right. \\
    u(-1, t) &= u(1, t) = 0
\end{aligned}\label{eqn:result:wave}
\end{equation}
up to time $T=0.5$. The network has $3$ intermediate layers, each of which has 100 neurons. The activation function is $\tanh$ except for the last layer. The network is trained for 100,000 epochs. Then we convert it into SNN. The results are shown in \ref{fig:result:pinn-wave}.

\begin{figure}[H]
    \centering
    \begin{subfigure}{0.245\linewidth}
        \includegraphics[width=\linewidth]{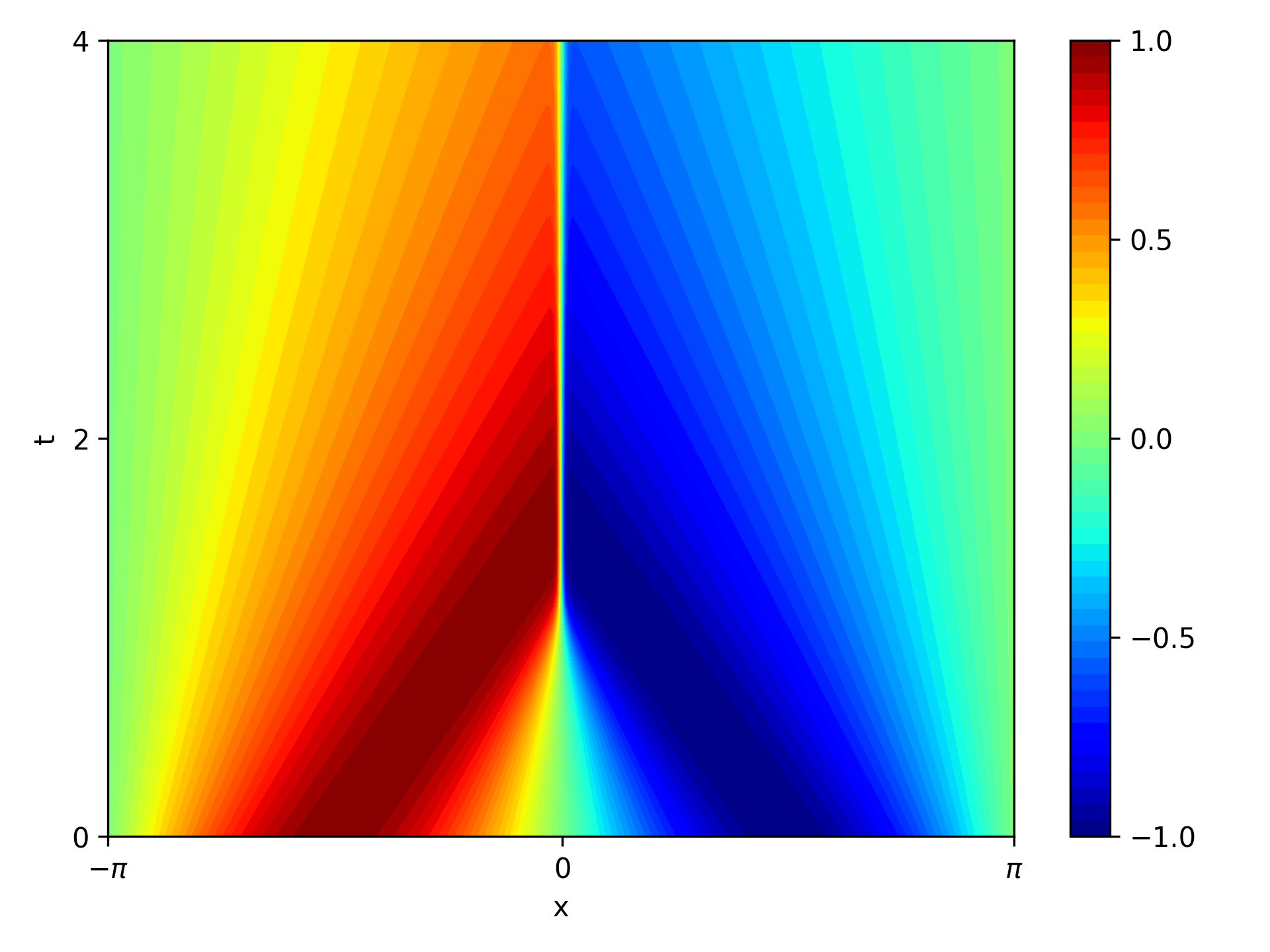}
        \caption{Reference solution}\label{fig:result:pinn-burgers:true}
    \end{subfigure}
    \begin{subfigure}{0.245\linewidth}
        \includegraphics[width=\linewidth]{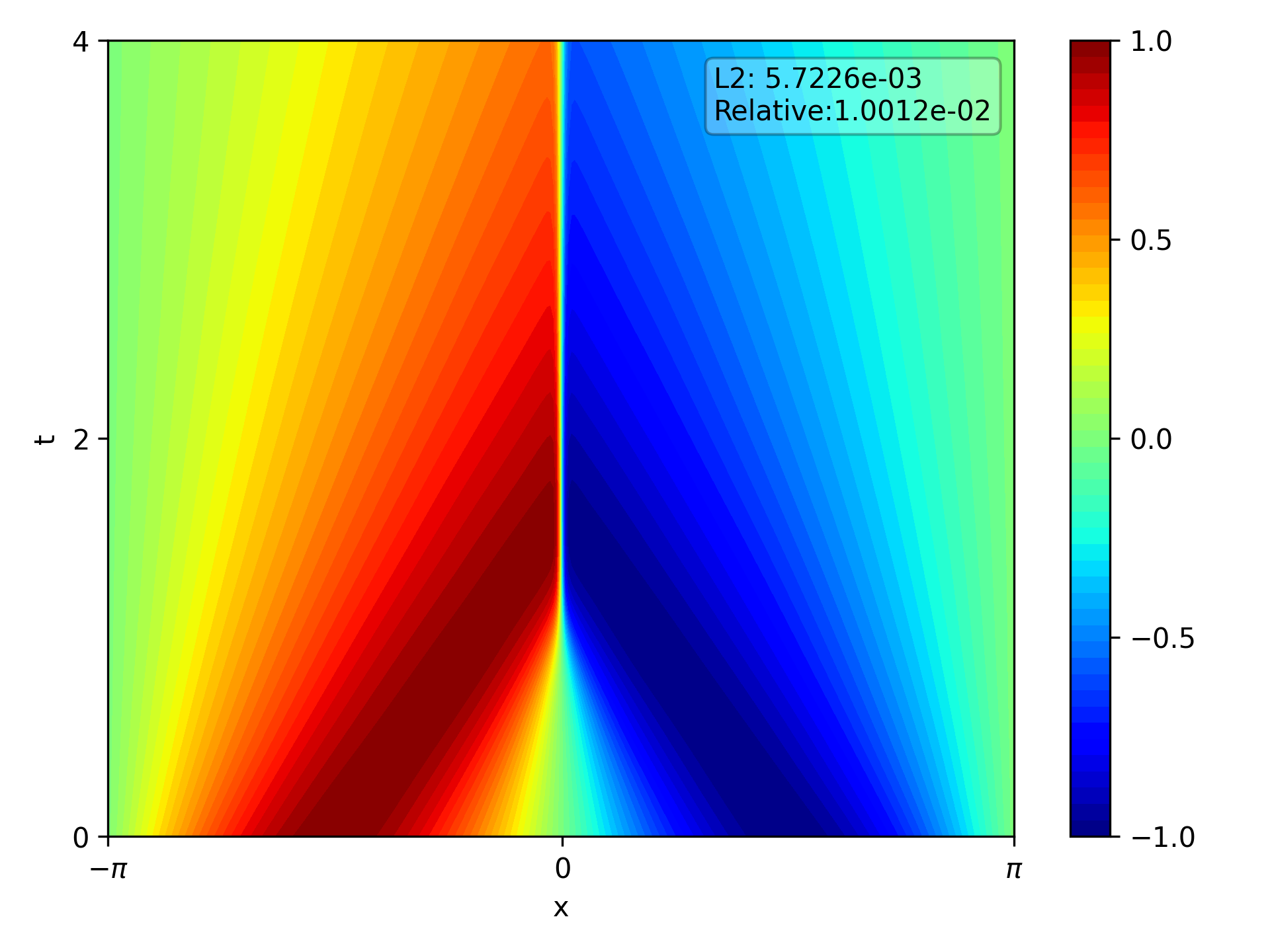}
        \caption{PINN result}\label{fig:result:pinn-burgers:ann}
    \end{subfigure}
    \begin{subfigure}{0.245\linewidth}
        \includegraphics[width=\linewidth]{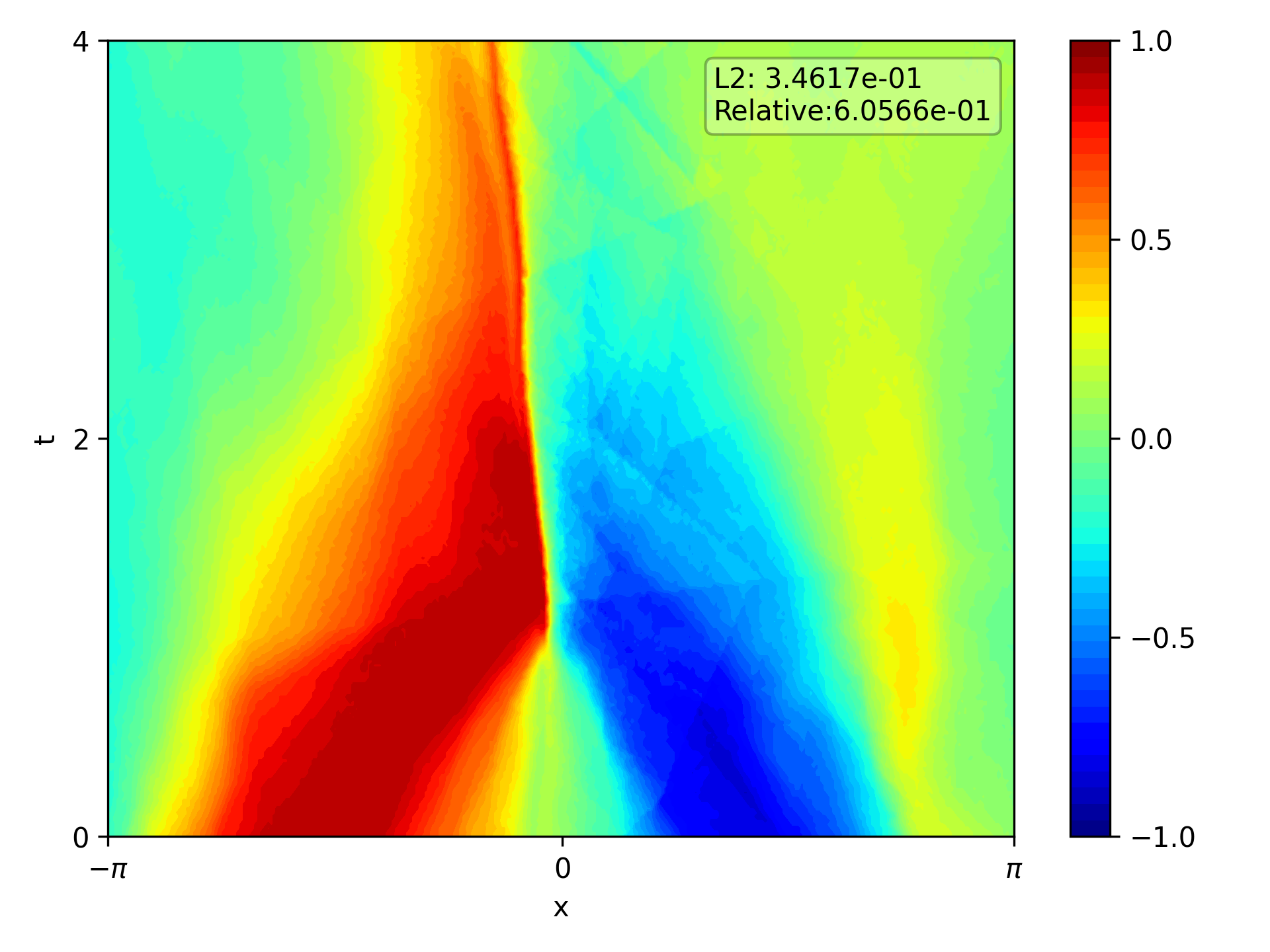}
        \caption{Conversion w/o calibration}\label{fig:result:pinn-burgers:snn_none}
    \end{subfigure}
    \begin{subfigure}{0.245\linewidth}
        \includegraphics[width=\linewidth]{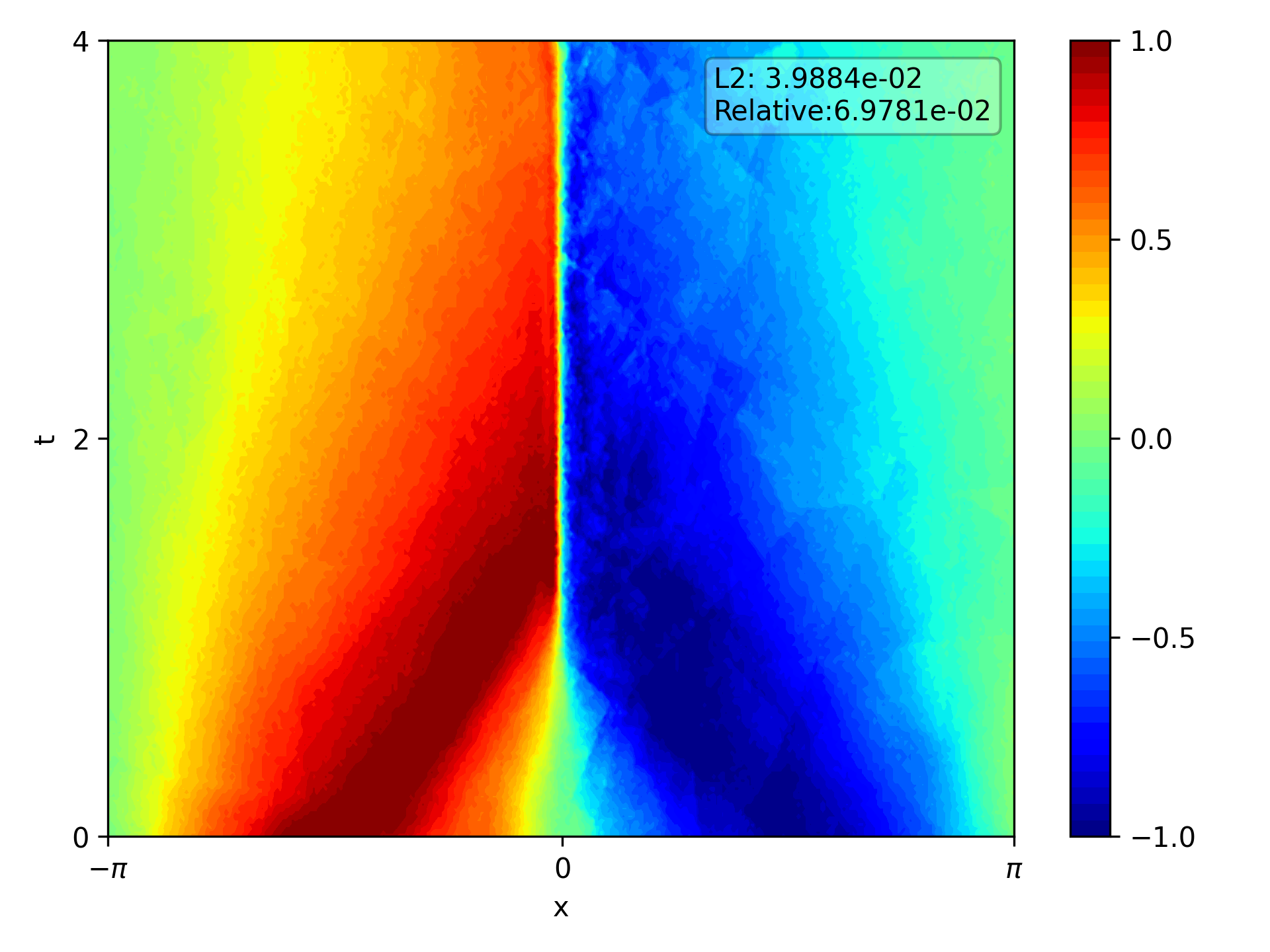}
        \caption{Conversion w/ calibration}\label{fig:result:pinn-burgers:snn_advanced}
    \end{subfigure}
    \caption{Burgers equation: The results of converting a PINN solving the viscous Burgers equation (\ref{eqn:result:burgers}). Figure \ref{fig:result:pinn-burgers:true} is the reference solution. Figure \ref{fig:result:pinn-burgers:ann} is the PINN result. Figure \ref{fig:result:pinn-burgers:snn_none} is the result of the SNN converted from the PINN without using calibration. Figure \ref{fig:result:pinn-burgers:snn_none} is the result of the SNN converted from the PINN using calibration.}\label{fig:result:pinn-burgers}
\end{figure}

\paragraph{Viscous Burgers equation} The Burgers equation is a prototype PDE representing nonlinear advection-diffusion occurs fluid mechanics. Here we solve the following problem of viscous Burgers equation:
\begin{equation}
\begin{aligned}
    \frac{\partial u}{\partial t} -\frac{\partial}{\partial x} (\frac{1}{2}u^2) &= \nu\frac{\partial^2 u}{\partial x^2},\quad (x, t)\in[0,2\pi]\times[0,4]\\
    u(x, 0) &= \sin(x)
\end{aligned}\label{eqn:result:burgers}
\end{equation}
with PINN. The network has $6$ intermediate layers, each of which has $40$ neurons. The activation function is $\tanh$ except for the last layer. The network is trained for 100,000 epochs. Then we convert it into SNN. The results are shown in \ref{fig:result:pinn-burgers}. 
Here we can find that conversion without calibration does not give correct position of steep gradient as the position is moving left (see Figure \ref{fig:result:pinn-burgers:snn_none}). But the conversion with calibration keeps the steep gradient position correct, which is important for physics.

Due to the discontinuous nature of SNN, the conversion results are not smooth. However, we can still apply some filters to smooth the outputs. For example, we apply FFT to the conversion results and remove the high frequencies, and the results are shown in Figure
\begin{figure}[htbp]
    \centering
    \begin{subfigure}{0.245\linewidth}
        \includegraphics[width=\linewidth]{figures/result/pinn/burgers-snn_advanced.png}
        \caption{Conversion w/ calibration}\label{fig:result:pinn-burgers:snn_advanced_original}
    \end{subfigure}
    \begin{subfigure}{0.245\linewidth}
        \includegraphics[width=\linewidth]{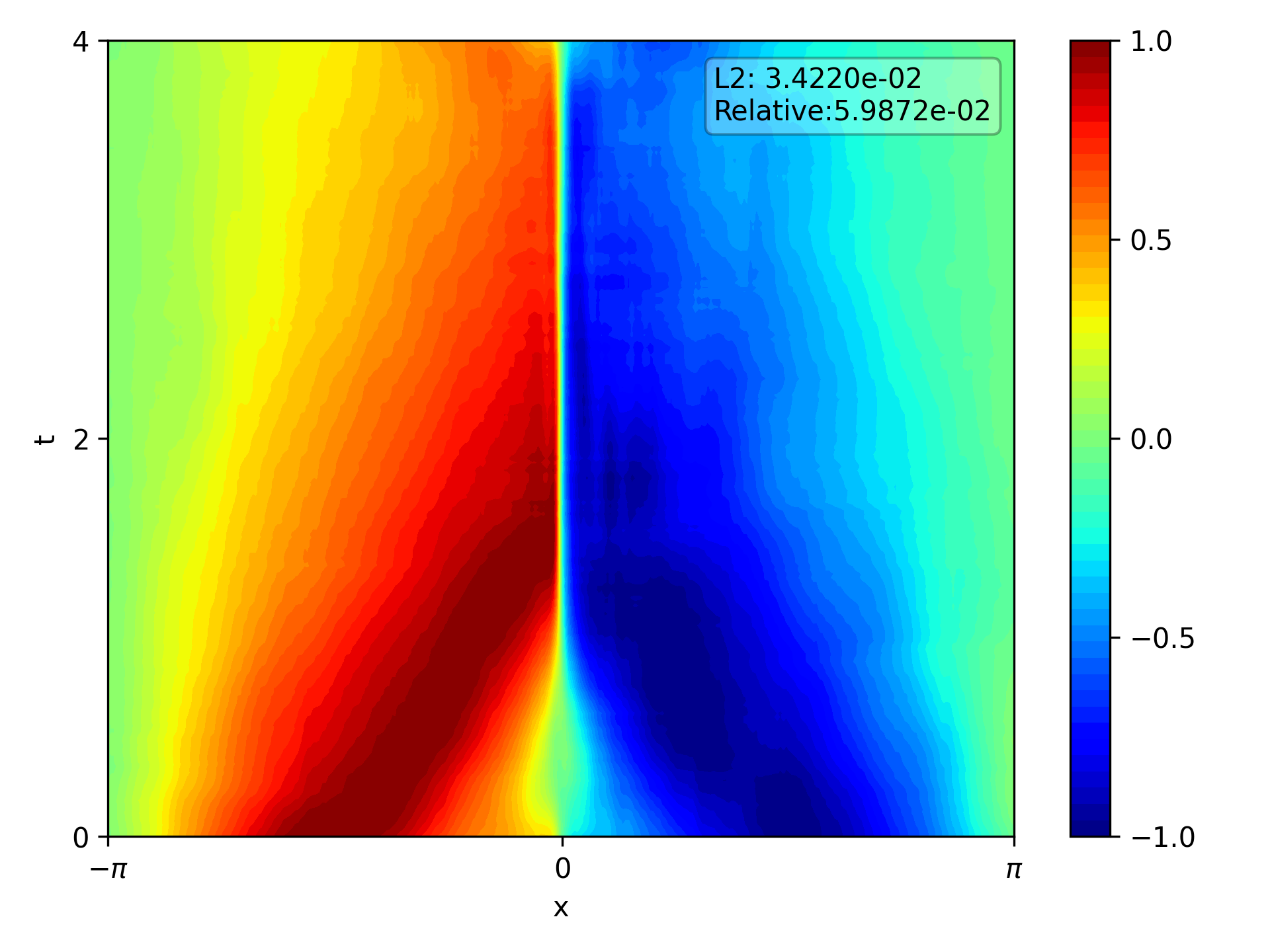}
        \caption{Conversion w/ calibration}\label{fig:result:pinn-burgers:snn_advanced_smooth}
    \end{subfigure}
    \caption{Figure \ref{fig:result:pinn-burgers:snn_advanced_original} is the original conversion result. Figure \ref{fig:result:pinn-burgers:snn_advanced_smooth} is the smoothed conversion result.}
\end{figure}
After the smoothing, conversion error becomes lower, hence can be adopted as the postprocessing procedure.

\subsection{Accelerating training with separable PINNs (SPINNs)}
\begin{figure}[htbp]
    \centering
    \begin{subfigure}{0.2455\linewidth}
        \includegraphics[width=\linewidth]{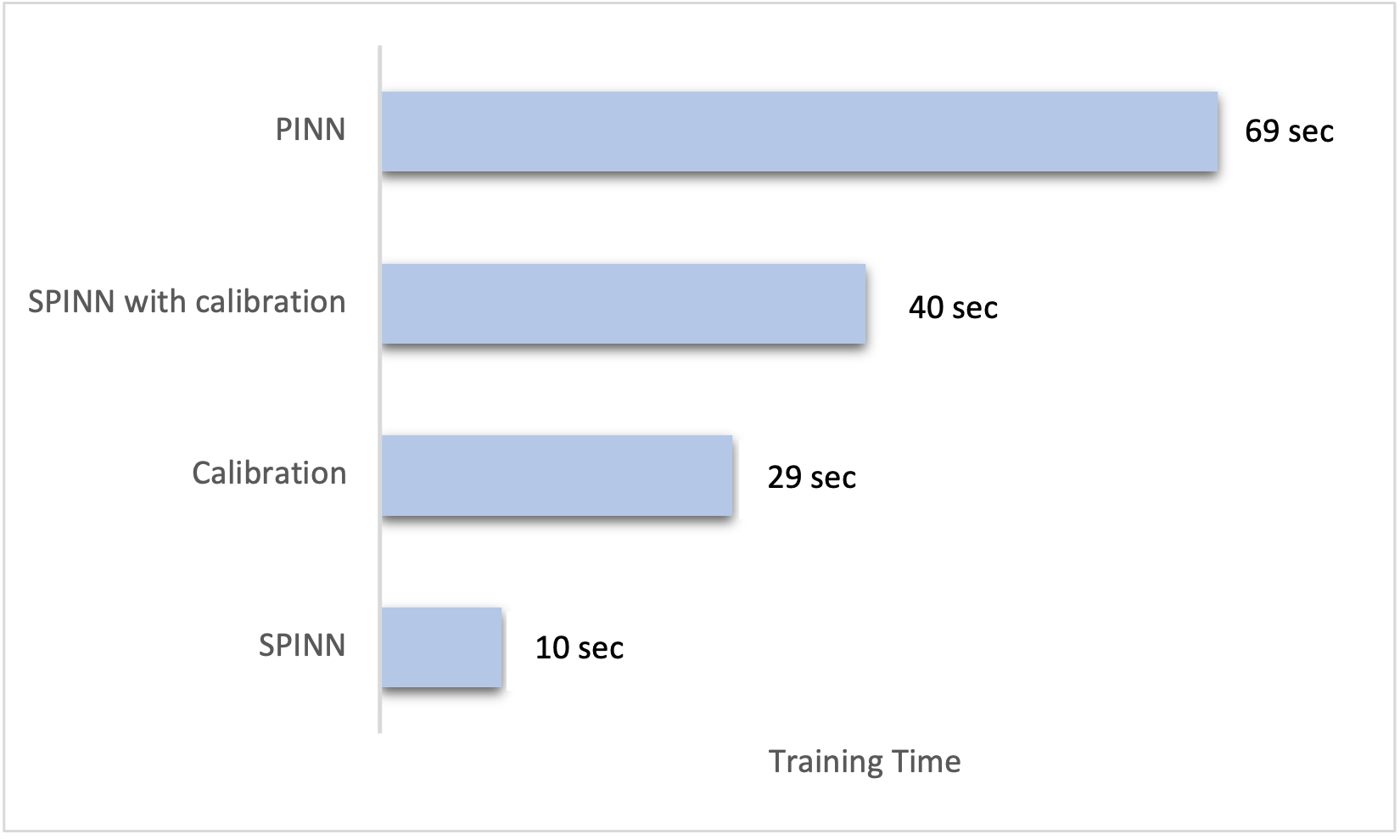}
        \caption{2D Burgers equation runtime}\label{fig:burgers}
    \end{subfigure}
    \begin{subfigure}{0.2455\linewidth}
        \includegraphics[width=\linewidth]{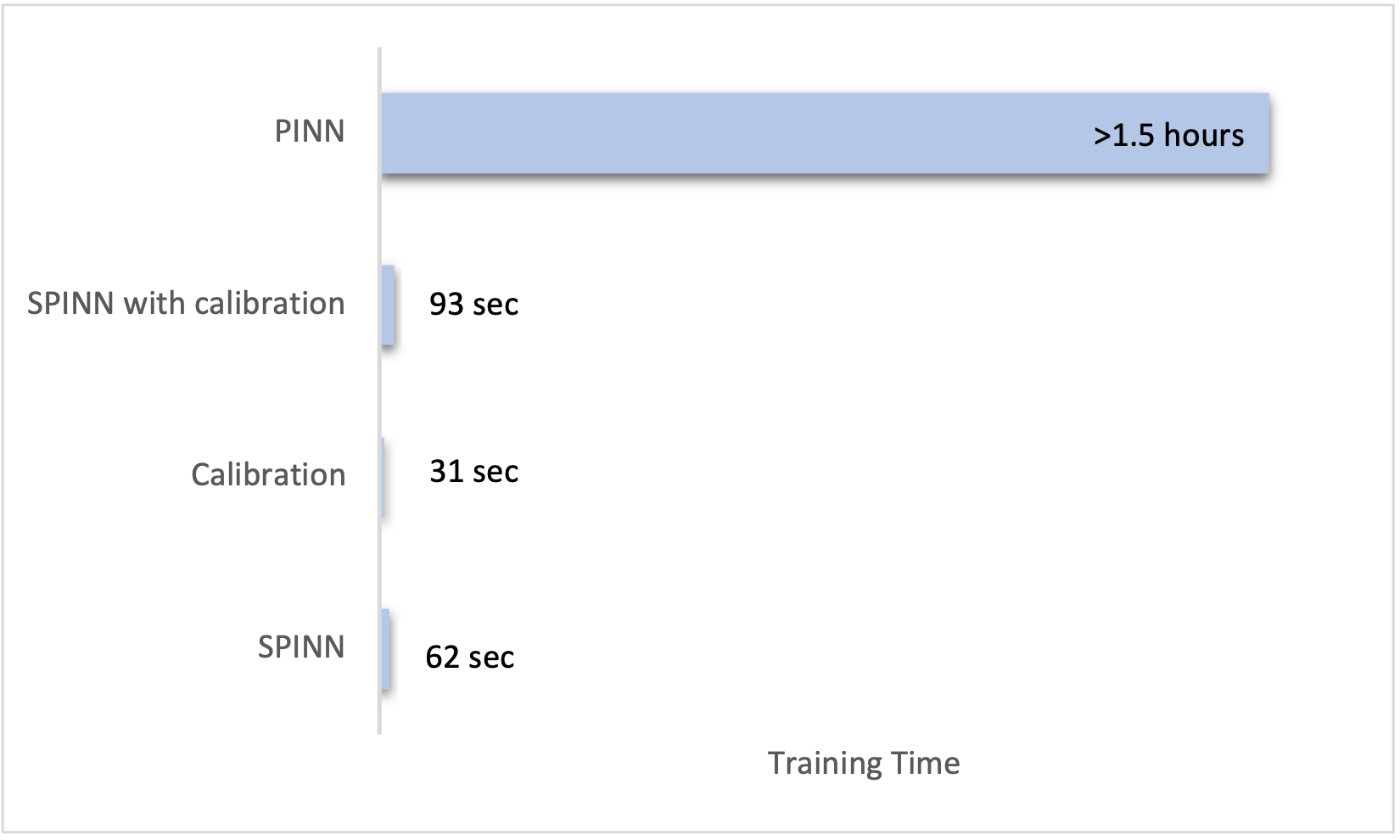}
        \caption{3D Beltrami flow runtime}\label{fig:result:NSbeltrami}
    \end{subfigure}
    \caption{The runtimes of the SPINN-SNN conversions solving the viscous Burgers equation (\ref{eqn:result:burgers}) and Beltrami flow (\ref{eqn:result:beltrami}). Figure \ref{fig:burgers} is the runtime for the 2D viscous Burgers equation. Figure \ref{fig:result:NSbeltrami} is the runtime for the 3D Beltrami flow.} \label{fig:result:runtime}
\end{figure}

A current limitation of PINN-SNN conversion is the computational resources it requires, especially for training the ANN. Herein, we highlight the effectiveness of using the separable physics-informed neural networks to spiking neural networks (SPINN-SNN) conversion. The SPINN-SNN conversion pipeline enhances the speed of operations and provides great computational efficiency compared to the direct application of PINN. We implement the SPINN-SNN conversion to address two PDEs: a two-dimensional viscous Burgers equation and a three-dimensional unsteady Beltrami flow. We provide a comparative analysis of the training time taken by standard PINN and the SPINN-SNN conversion pipeline. Experimental results reveal that the application of SPINN-SNN conversion greatly enhances the speed of the training, particularly when solving high dimensional PDEs. Figure \ref{fig:result:runtime} displays a runtime comparison between SPINN-SNN conversion and PINN, applied to two and three-dimensional problems. When addressing the two-dimensional Burgers equation, the SPINN-SNN conversion process is approximately 1.7 times faster than PINN. The superiority of SPINN-SNN conversion becomes more apparent while solving the three-dimensional Beltrami flow problem, which is over 60 times faster than PINN. Notably, the time necessary for the SNN calibration remains relatively constant, regardless of the problem dimensionality. This indicates that the benefits of the SPINN-SNN conversion pipeline become increasingly prominent with the rise in the dimensionality of the problem.

\begin{figure}[htbp]
    \centering
    \begin{subfigure}{0.245\linewidth}
        \includegraphics[width=\linewidth]{figures/result/pinn/burgers-true.png}
        \caption{Reference solution}\label{fig:result:spinn-burgers:true}
    \end{subfigure}
    \begin{subfigure}{0.245\linewidth}
        \includegraphics[width=\linewidth]{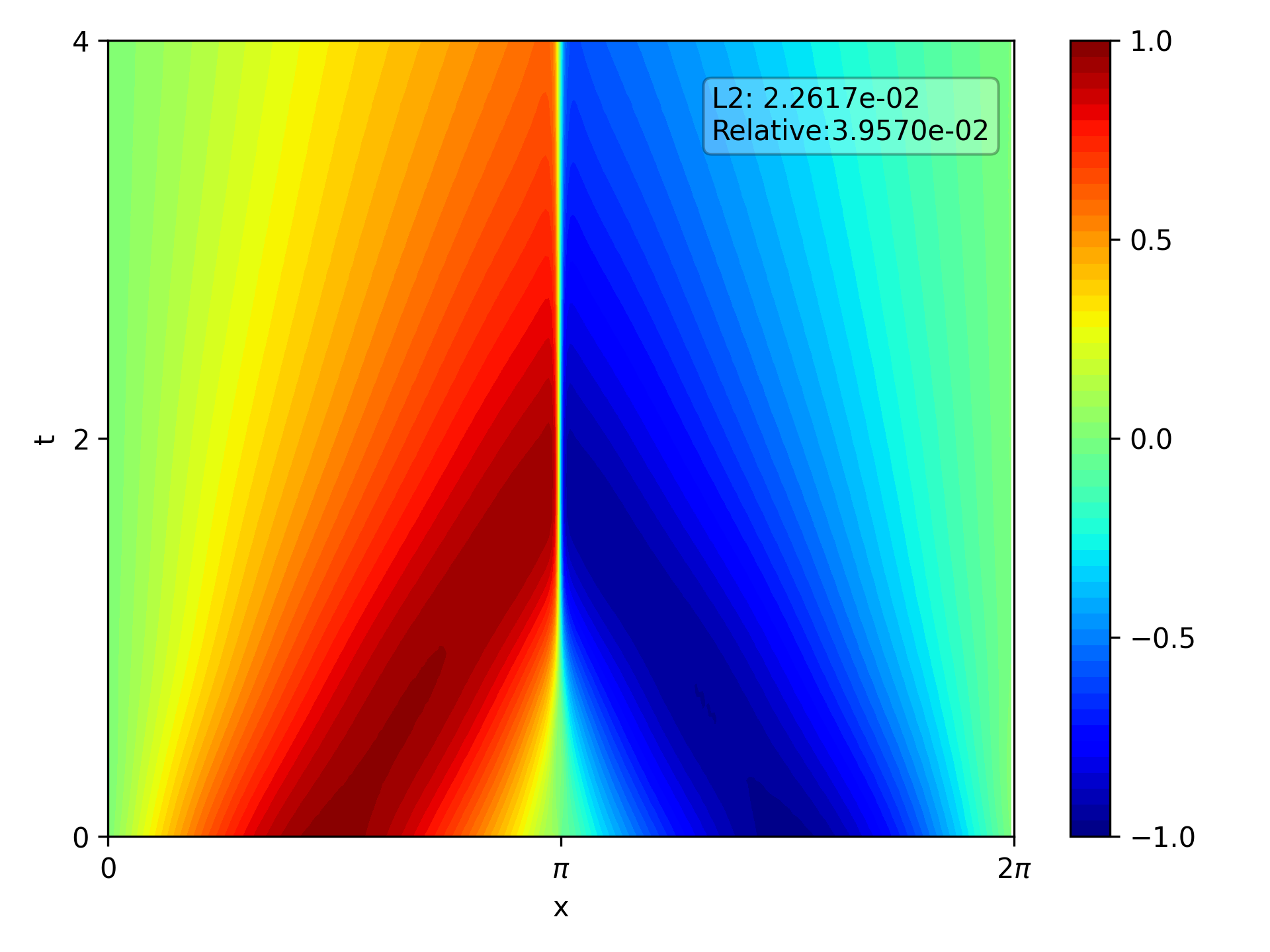}
        \caption{Separable PINN result}\label{fig:result:spinn-burgers:ann}
    \end{subfigure}
    \begin{subfigure}{0.245\linewidth}
        \includegraphics[width=\linewidth]{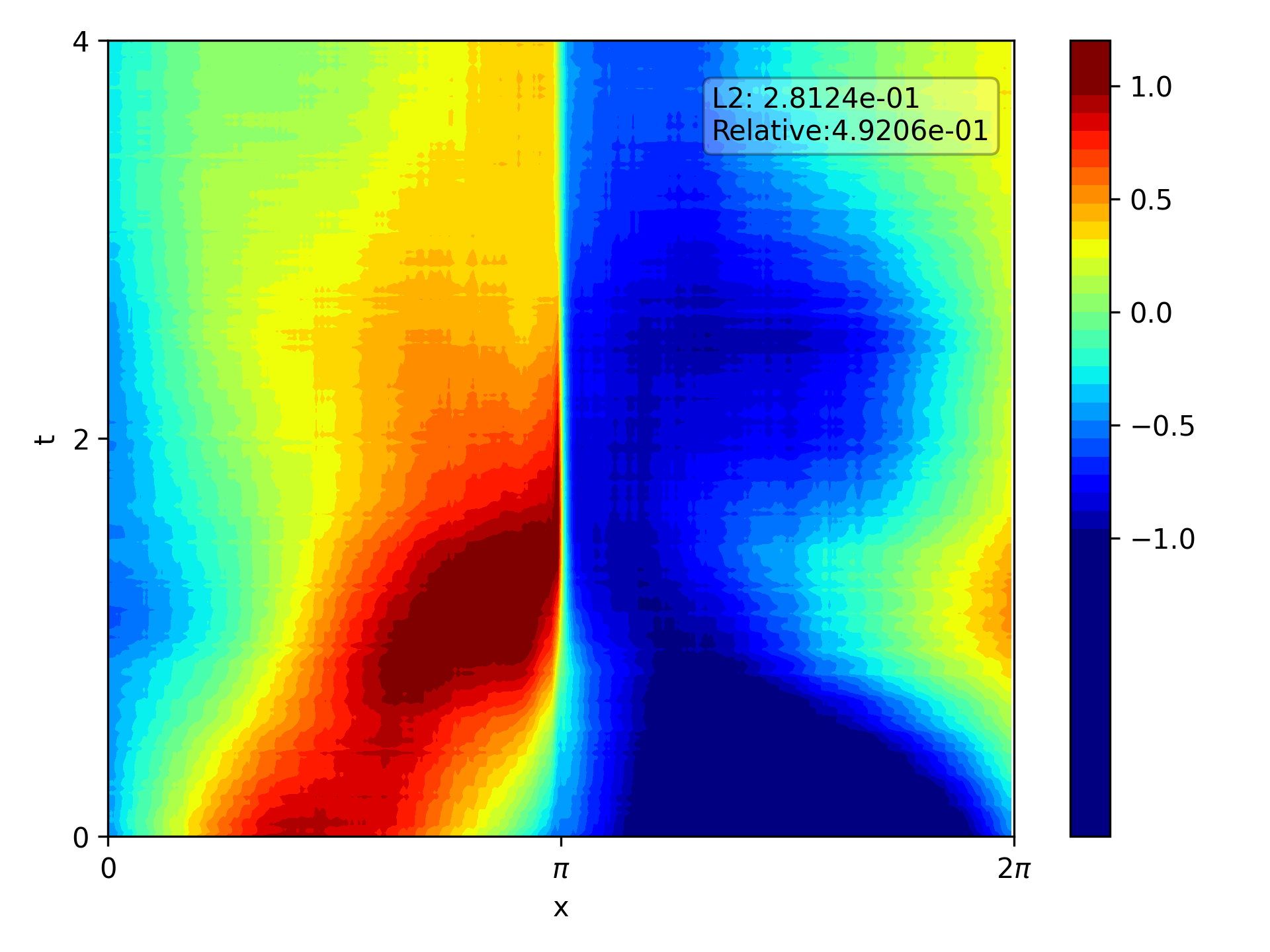}
        \caption{Conversion w/o calibration}\label{fig:result:spinn-burgers:snn_none}
    \end{subfigure}
    \begin{subfigure}{0.245\linewidth}
        \includegraphics[width=\linewidth]{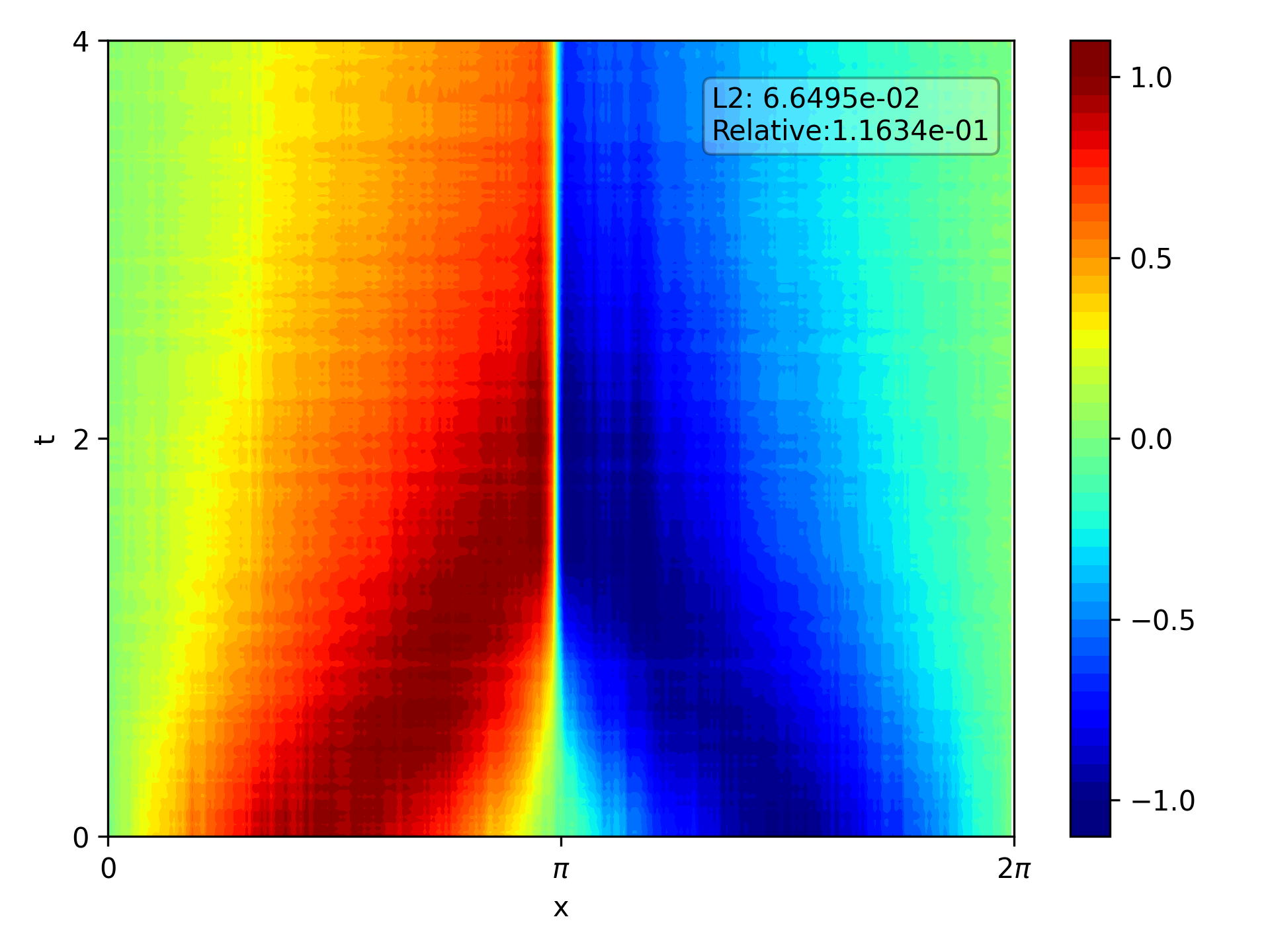}
        \caption{Conversion w/ calibration}\label{fig:result:spinn-burgers:snn_advanced}
    \end{subfigure}
    \caption{Burgers equation with Separable PINN (SPINN): The results of converting a Separable PINN solving the viscous Burgers equation (\ref{eqn:result:burgers}). Figure \ref{fig:result:spinn-burgers:true} is the reference solution. Figure \ref{fig:result:spinn-burgers:ann} is the Separable PINN result. Figure \ref{fig:result:spinn-burgers:snn_none} is the result of the SNN converted from the Separable PINN without using calibration. Figure \ref{fig:result:spinn-burgers:snn_none} is the result of the SNN converted from the Separable PINN using calibration.}\label{fig:result:spinn-burgers}
\end{figure}

\paragraph{Viscous Burgers equation} In order to conduct a fair comparison between the PINN-SNN and SPINN-SNN conversions, the setup of the viscous Burgers equation is kept consistent with that described in Equation \ref{eqn:result:burgers}, and the same set of hyperparameters is utilized. Figure \ref{fig:result:spinn-burgers} presents the results of converting a SPINN to solve the Burgers’ equation. The SPINN has individual subnetworks for each independent variable, $x$ and $t$. Each of these subnetworks comprises three intermediate layers, each layer containing 40 neurons, and employs the tanh activation function, except  in the last layer. As depicted in Figure \ref{fig:result:spinn-burgers}, the conversion from SPINN provides an accuracy level comparable to that of PINN. An SNN converted with calibration achieves a significantly smaller error than one converted without calibration.

\paragraph{Beltrami Flow}

The Navier-Stokes equations are fundamental in fluid mechanics as they mathematically represent the conservation of momentum in fluid systems, and there has been significant advancement in solving Navier-Stokes flow problems using scientific machine learning methods \cite{jin2021nsfnets,ns_eqn1,ns_eqn2}. The Navier-Stokes equations can be presented in two forms: the velocity-pressure (VP) form and the vorticity-velocity (VV) form. The incompressible Navier-Stokes equations, in their VP form, are as follows:

\begin{equation}
\begin{aligned}
    \frac{\partial{\mathbf{u}}}{\partial t} + (\mathbf{u} \cdot \nabla )\mathbf{u} &= -\nabla p + \frac{1}{Re}\nabla^2 \mathbf{u}\\
    \nabla \cdot \mathbf{u} &= 0
\end{aligned}\label{eqn:result:beltrami}
\end{equation}

We chose the spatial domain to be $\Omega \in [-1,1]^2$ and time interval to be $\Gamma \in [0,1]$. Here, $t$ is non-dimensional time, $\mathbf{u}(\mathbf{x},t) = [u,v]^T$ the non-dimensional velocity in the $(x,y)$-directions, $p$ the non-dimensional pressure, and the Reynolds number $Re = \frac{U_{ref} D_{ref}}{v}$ is defined by characteristic length ($D_{ref}$), reference velocity ($U_{ref}$), and kinematic viscosity ($v$). In this example, we simulate a three-dimensional unsteady laminar Beltrami flow where $Re = 1$. The analytical solution of the Beltrami flow is \cite{kim1985NSsolution}:

\begin{equation}
\begin{aligned}
    u(x,y,t) &= -\cos x \sin y\text{ }e^{-2t}\\
    v(x,y,t) &= \sin x \cos y\text{ }e^{-2t}\\
    p(x,y,t) &= -\frac{1}{4} (\cos 2x + \cos 2y)\text{ }e^{-4t}
\end{aligned}
\end{equation}

\begin{figure}[htbp]
    \centering
    \includegraphics[scale = 0.65]{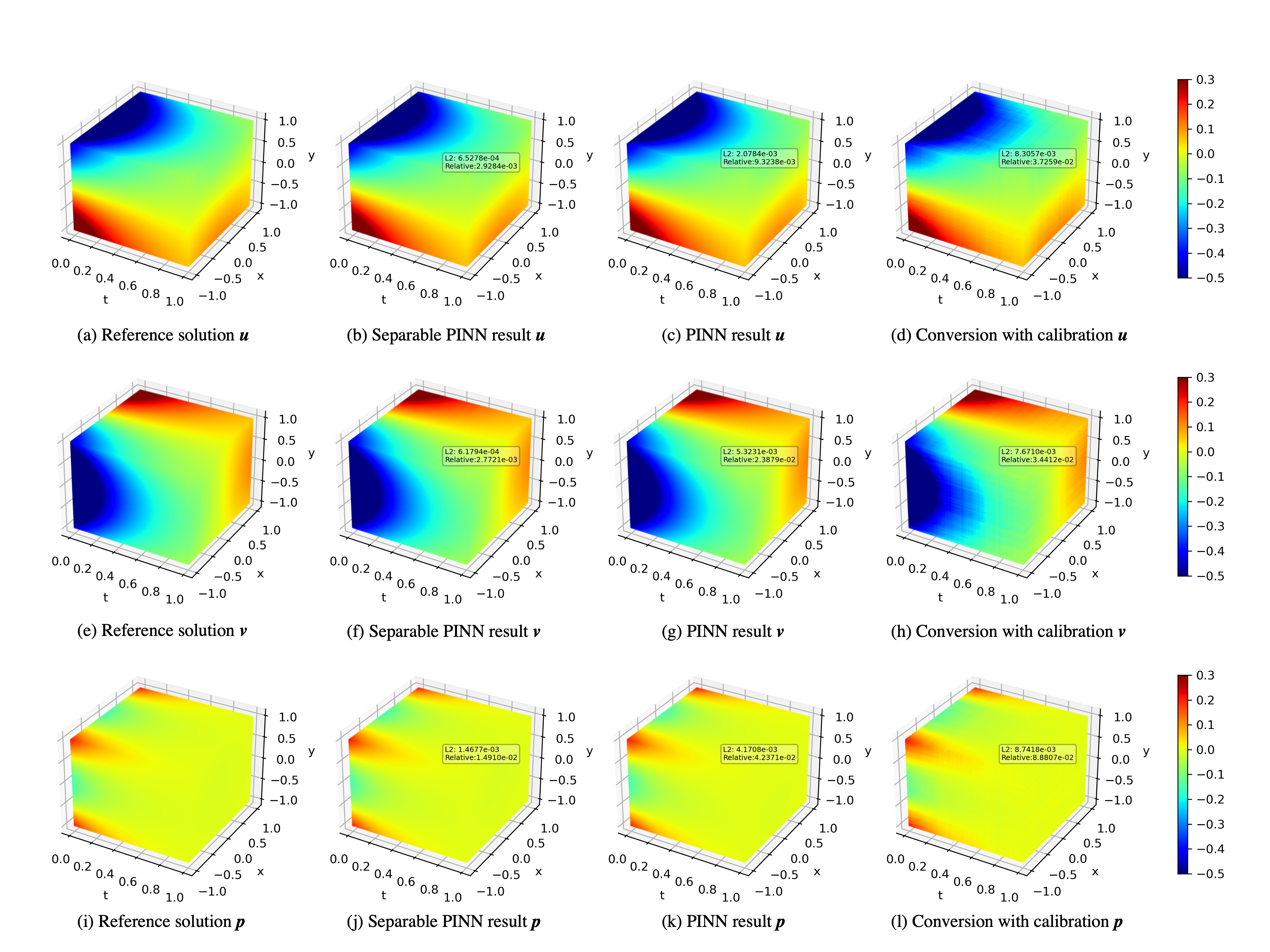}
    \caption{Beltrami flow with Separable PINN (SPINN): The results of converting a SPINN solving the Beltrami flow problem(\ref{eqn:result:burgers}). Figure \ref{fig:result:spinn-beltrami}(a,e,i) are the reference solutions for $\mathbf{u},\mathbf{v},\mathbf{p}$ respectively. Figure \ref{fig:result:spinn-beltrami}(b,f,j) are the SPINN results. Figure \ref{fig:result:spinn-beltrami}(c,g,k) are the PINN results. Figure \ref{fig:result:spinn-beltrami}(d,h,i) are the results of the SNN converted from the SPINN with calibration.}
    \label{fig:result:spinn-beltrami}
\end{figure}

The boundary and initial conditions are extracted from this exact solution. The PINN network comprises 4 intermediate layers, with each one containing 128 neurons. The activation function is tanh applied taoi all layers except the final layer. The network is trained for 20,000 epochs before being converted into an SNN. The SPINN consists of separate subnetworks for each independent variable, u, v, and p. Every subnetwork has 2 intermediate layers, with each layer consisting of 50 neurons. They all utilize the tanh activation function, apart from the last layer. Figure \ref{fig:result:spinn-beltrami} illustrates that the SPINN conversion offers similar accuracy compared to PINN. The error for pressure is slightly more than the velocity errors in both the $x$ and $y$ axes. Nevertheless, the SNN, once calibrated and converted using SPINN, attains good accuracy and notable speed improvements in comparison to PINN.

\subsection{Firing rates of the converted SNN}
To demonstrate the potential efficiency of converting the ANN to the SNN, we computed the spiking rates of different equations. The spiking rate is defined as the ratio of non-zero values in the output of each layer. Prior works have suggested that SNNs with lower spiking rates will translate to energy efficient impelmentations on neurmorphic hardware. 
The results are shown in Table 1. We observe $<0.5$ spiking rate in most cases demonstrating that SNNs only expend $<50\%$ of their network computations.
%

\begin{table}[H]
\caption{The spiking rate of SNNs for different equations.}
\centering
\begin{tabular}{c|cc}
\toprule
Equations & Number of parameters & Spiking rate \\
\midrule
Poisson & 20601 & 0.3727 \\
Diffusion-reaction & 20601 & 0.2879 \\
Wave & 20601 & 0.5721 \\
Burgers & 8361 & 0.7253 \\
Burgers (Separable PINN) & 15500 & 0.2831 \\
N-S (Beltrami flow, Separable PINN) & 30900 & 0.1754 \\
\bottomrule
\end{tabular}
\end{table} 

\begin{table}[H]
\caption{The $L^2$ and relative $L^2$ error of converted SNN for different equations.}
\centering
\begin{tabular}{c|cc}
\toprule
Equations & $L^2$ error & Relative $L^2$ error \\
\midrule
Poisson & $7.8508\times10^{-3}$ & $4.7095\times10^{-2}$\\
Diffusion-reaction & $2.8766\times10^{-2}$ & $6.3267\times10^{-2}$ \\
Wave & $3.8965\times10^{-2}$ & $6.5308\times10^{-2}$ \\
Burgers & $3.9884\times10^{-2}$ & $6.9781\times10^{-2}$ \\ 
Burgers (Separable PINN) & $6.6495\times10^{-2}$ & $1.1634\times10^{-1}$ \\
N-S (Beltrami flow, Separable PINN) & $8.2512\times10^{-3}$ & $4.3273\times10^{-2}$ \\
\bottomrule
\end{tabular}
\end{table} 

\section{Conclusion}
We have successfully extended the SNN calibration method proposed in \cite{li2022calibration} to a more generalized class of activation functions beyond ReLU. The original proof relied on the specific property of ReLU's second-order derivative being zero, but our approach relaxed this constraint by incorporating the training loss as an additional term in the bound. We demonstrated the effectiveness of our method through various examples, including PINN and Separable PINN \cite{cho2022separable} (a variant of PINN), where the activation functions are not ReLU. The results demonstrated that our approach achieved good accuracy with low spike rates, making it a promising and energy-efficient solution for scientific machine learning. By enabling the conversion of a wider range of neural networks to SNNs, this method opens up new possibilities for harnessing the temporal processing capabilities and computational efficiency of SNNs in various scientific applications. The proposed approach contributes to advancing the field of spiking neural networks and their potential for practical real-world implementations in edge computing. Future research can explore further extensions of the calibration technique to other types of activation functions and investigate its performance on more complex neural network architectures.

\section{Acknowledgements}
This work was supported in part by the DOE SEA-CROGS project (DE-SC0023191), the ONR Vannevar Bush Faculty Fellowship (N00014-22-1-2795), CoCoSys- a JUMP2.0
center sponsored by DARPA and SRC, and the National Science Foundation CAREER
Award.

\bibliographystyle{unsrt}
\bibliography{references}

\begin{thebibliography}{10}

\bibitem{pinn_fluid}
Shengze Cai, Zhiping Mao, Zhicheng Wang, Minglang Yin, and George~Em
  Karniadakis.
\newblock Physics-informed neural networks ({PINNs}) for fluid mechanics: A
  review.
\newblock {\em Acta Mechanica Sinica}, pages 1--12, 2022.

\bibitem{ml_physics}
Qianying Cao, Somdatta Goswami, George~Em Karniadakis, and Souvik Chakraborty.
\newblock Deep neural operators can predict the real-time response of floating
  offshore structures under irregular waves.
\newblock {\em arXiv preprint arXiv:2302.06667}, 2023.

\bibitem{JIN2020SympNets}
Pengzhan Jin, Zhen Zhang, Aiqing Zhu, Yifa Tang, and George~Em Karniadakis.
\newblock Sympnets: Intrinsic structure-preserving symplectic networks for
  identifying hamiltonian systems.
\newblock {\em Neural Networks}, 132:166--179, 2020.

\bibitem{Zhang2022HINTS}
Enrui Zhang, Adar Kahana, Eli Turkel, Rishikesh Ranade, Jay Pathak, and
  George~Em Karniadakis.
\newblock A hybrid iterative numerical transferable solver (hints) for pdes
  based on deep operator network and relaxation methods, 2022.

\bibitem{ml_chemistry1}
Somdatta Goswami, Ameya~D Jagtap, Hessam Babaee, Bryan~T Susi, and George~Em
  Karniadakis.
\newblock Learning stiff chemical kinetics using extended deep neural
  operators.
\newblock {\em arXiv preprint arXiv:2302.12645}, 2023.

\bibitem{ml_chemistry2}
Qianying Cao, Somdatta Goswami, and George~Em Karniadakis.
\newblock Lno: Laplace neural operator for solving differential equations.
\newblock {\em arXiv preprint arXiv:2303.10528}, 2023.

\bibitem{pinn_sysbio}
Mitchell Daneker, Zhen Zhang, George~Em Karniadakis, and Lu~Lu.
\newblock Systems biology: Identifiability analysis and parameter
  identification via systems-biology informed neural networks.
\newblock {\em arXiv preprint arXiv:2202.01723}, 2022.

\bibitem{ml_biology}
Juan~Diego Toscano, Christian Zuniga-Navarrete, Wilson David~Jo Siu,
  Luis~Javier Segura, and Hongyue Sun.
\newblock Teeth mold point cloud completion via data augmentation and hybrid
  rl-gan.
\newblock {\em Journal of Computing and Information Science in Engineering},
  23(4):041008, 2023.

\bibitem{Zhang2022AOSLO}
Qian Zhang, Konstantina Sampani, Mengjia Xu, Shengze Cai, Yixiang Deng, He~Li,
  Jennifer~K. Sun, and George~Em Karniadakis.
\newblock {AOSLO-net: A Deep Learning-Based Method for Automatic Segmentation
  of Retinal Microaneurysms From Adaptive Optics Scanning Laser Ophthalmoscopy
  Images}.
\newblock {\em Translational Vision Science \& Technology}, 11(8):7--7, 08
  2022.

\bibitem{geologics_ccs}
Zhongyi Jiang, Min Zhu, Dongzhuo Li, Qiuzi Li, Yanhua~O Yuan, and Lu~Lu.
\newblock Fourier-mionet: Fourier-enhanced multiple-input neural operators for
  multiphase modeling of geological carbon sequestration.
\newblock {\em arXiv preprint arXiv:2303.04778}, 2023.

\bibitem{geologics_fwi}
Min Zhu, Shihang Feng, Youzuo Lin, and Lu~Lu.
\newblock Fourier-deeponet: Fourier-enhanced deep operator networks for full
  waveform inversion with improved accuracy, generalizability, and robustness.
\newblock {\em arXiv preprint arXiv:2305.17289}, 2023.

\bibitem{Zhang2021PLOS}
Sheng Zhang, Joan Ponce, Zhen Zhang, Guang Lin, and George Karniadakis.
\newblock An integrated framework for building trustworthy data-driven
  epidemiological models: Application to the covid-19 outbreak in new york
  city.
\newblock {\em PLOS Computational Biology}, 17(9):1--29, 09 2021.

\bibitem{Kharazmi2021}
Ehsan Kharazmi, Min Cai, Xiaoning Zheng, Zhen Zhang, Guang Lin, and George~Em
  Karniadakis.
\newblock Identifiability and predictability of integer- and fractional-order
  epidemiological models using physics-informed neural networks.
\newblock {\em Nature Computational Science}, 1(11):744--753, Nov 2021.

\bibitem{raissi2019physics}
Maziar Raissi, Paris Perdikaris, and George~E Karniadakis.
\newblock Physics-informed neural networks: A deep learning framework for
  solving forward and inverse problems involving nonlinear partial differential
  equations.
\newblock {\em Journal of Computational physics}, 378:686--707, 2019.

\bibitem{lu2021deepxde}
Lu~Lu, Xuhui Meng, Zhiping Mao, and George~Em Karniadakis.
\newblock Deepxde: A deep learning library for solving differential equations.
\newblock {\em SIAM review}, 63(1):208--228, 2021.

\bibitem{pinn_sampling}
Chenxi Wu, Min Zhu, Qinyang Tan, Yadhu Kartha, and Lu~Lu.
\newblock A comprehensive study of non-adaptive and residual-based adaptive
  sampling for physics-informed neural networks.
\newblock {\em Computer Methods in Applied Mechanics and Engineering},
  403:115671, 2023.

\bibitem{pinn_attention}
Sokratis~J Anagnostopoulos, Juan~Diego Toscano, Nikolaos Stergiopulos, and
  George~Em Karniadakis.
\newblock Residual-based attention and connection to information bottleneck
  theory in pinns.
\newblock {\em arXiv preprint arXiv:2307.00379}, 2023.

\bibitem{g_pinn}
Jeremy Yu, Lu~Lu, Xuhui Meng, and George~Em Karniadakis.
\newblock Gradient-enhanced physics-informed neural networks for forward and
  inverse pde problems.
\newblock {\em Computer Methods in Applied Mechanics and Engineering},
  393:114823, 2022.

\bibitem{Zou2023MHNN}
Zongren Zou and George~Em Karniadakis.
\newblock L-hydra: Multi-head physics-informed neural networks, 2023.

\bibitem{pinn_finance}
Benjamin Fan, Edward Qiao, Anran Jiao, Zhouzhou Gu, Wenhao Li, and Lu~Lu.
\newblock Deep learning for solving and estimating dynamic macro-finance
  models.
\newblock {\em arXiv preprint arXiv:2305.09783}, 2023.

\bibitem{pinn_geometry}
Enrui Zhang, Ming Dao, George~Em Karniadakis, and Subra Suresh.
\newblock Analyses of internal structures and defects in materials using
  physics-informed neural networks.
\newblock {\em Science Advances}, 8(7):eabk0644, 2022.

\bibitem{pinn_hfm}
Maziar Raissi, Alireza Yazdani, and George~Em Karniadakis.
\newblock Hidden fluid mechanics: Learning velocity and pressure fields from
  flow visualizations.
\newblock {\em Science}, 367(6481):1026--1030, 2020.

\bibitem{pinn_material}
Enrui Zhang, Minglang Yin, and George~Em Karniadakis.
\newblock Physics-informed neural networks for nonhomogeneous material
  identification in elasticity imaging.
\newblock {\em arXiv preprint arXiv:2009.04525}, 2020.

\bibitem{pinn_optics}
Yuyao Chen, Lu~Lu, George~Em Karniadakis, and Luca Dal~Negro.
\newblock Physics-informed neural networks for inverse problems in nano-optics
  and metamaterials.
\newblock {\em Optics Express}, 28(8):11618--11633, 2020.

\bibitem{Zou2022NeuralUQ}
Zongren Zou, Xuhui Meng, Apostolos~F Psaros, and George~Em Karniadakis.
\newblock Neuraluq: A comprehensive library for uncertainty quantification in
  neural differential equations and operators, 2022.

\bibitem{PSAROS2023UQ}
Apostolos~F. Psaros, Xuhui Meng, Zongren Zou, Ling Guo, and George~Em
  Karniadakis.
\newblock Uncertainty quantification in scientific machine learning: Methods,
  metrics, and comparisons.
\newblock {\em Journal of Computational Physics}, 477:111902, 2023.

\bibitem{massa2020efficient}
Riccardo Massa, Alberto Marchisio, Maurizio Martina, and Muhammad Shafique.
\newblock An efficient spiking neural network for recognizing gestures with a
  dvs camera on the loihi neuromorphic processor.
\newblock In {\em 2020 International Joint Conference on Neural Networks
  (IJCNN)}, pages 1--9. IEEE, 2020.

\bibitem{kim2022rate}
Youngeun Kim, Hyoungseob Park, Abhishek Moitra, Abhiroop Bhattacharjee,
  Yeshwanth Venkatesha, and Priyadarshini Panda.
\newblock Rate coding or direct coding: Which one is better for accurate,
  robust, and energy-efficient spiking neural networks?
\newblock In {\em ICASSP 2022-2022 IEEE International Conference on Acoustics,
  Speech and Signal Processing (ICASSP)}, pages 71--75. IEEE, 2022.

\bibitem{bouvier2019spiking}
Maxence Bouvier, Alexandre Valentian, Thomas Mesquida, Francois Rummens, Marina
  Reyboz, Elisa Vianello, and Edith Beigne.
\newblock Spiking neural networks hardware implementations and challenges: A
  survey.
\newblock {\em ACM Journal on Emerging Technologies in Computing Systems
  (JETC)}, 15(2):1--35, 2019.

\bibitem{roy2019towards}
Kaushik Roy, Akhilesh Jaiswal, and Priyadarshini Panda.
\newblock Towards spike-based machine intelligence with neuromorphic computing.
\newblock {\em Nature}, 575(7784):607--617, 2019.

\bibitem{davies2018loihi}
Mike Davies, Narayan Srinivasa, Tsung-Han Lin, Gautham Chinya, Yongqiang Cao,
  Sri~Harsha Choday, Georgios Dimou, Prasad Joshi, Nabil Imam, Shweta Jain,
  et~al.
\newblock Loihi: A neuromorphic manycore processor with on-chip learning.
\newblock {\em Ieee Micro}, 38(1):82--99, 2018.

\bibitem{orchard2021efficient}
Garrick Orchard, E~Paxon Frady, Daniel Ben~Dayan Rubin, Sophia Sanborn,
  Sumit~Bam Shrestha, Friedrich~T Sommer, and Mike Davies.
\newblock Efficient neuromorphic signal processing with loihi 2.
\newblock In {\em 2021 IEEE Workshop on Signal Processing Systems (SiPS)},
  pages 254--259. IEEE, 2021.

\bibitem{Kahana2022SpikingNO}
Adar Kahana, Qian Zhang, Leonard Gleyzer, and George~Em Karniadakis.
\newblock Spiking neural operators for scientific machine learning, 2022.

\bibitem{Zhang2022SMS}
Qian Zhang, Adar Kahana, George~Em Karniadakis, and Panos Stinis.
\newblock Sms: Spiking marching scheme for efficient long time integration of
  differential equations, 2022.

\bibitem{nageswaran2009computing}
Jayram~Moorkanikara Nageswaran, Nikil Dutt, Yingxue Wang, and Tobi Delbrueck.
\newblock Computing spike-based convolutions on gpus.
\newblock In {\em 2009 IEEE International Symposium on Circuits and Systems
  (ISCAS)}, pages 1917--1920. IEEE, 2009.

\bibitem{bienenstock1982theory}
Elie~L Bienenstock, Leon~N Cooper, and Paul~W Munro.
\newblock Theory for the development of neuron selectivity: orientation
  specificity and binocular interaction in visual cortex.
\newblock {\em Journal of Neuroscience}, 2(1):32--48, 1982.

\bibitem{hodgkin1952measurement}
Alan~L Hodgkin, Andrew~F Huxley, and Bernard Katz.
\newblock Measurement of current-voltage relations in the membrane of the giant
  axon of loligo.
\newblock {\em The Journal of physiology}, 116(4):424, 1952.

\bibitem{izhikevich2007dynamical}
Eugene~M Izhikevich.
\newblock {\em Dynamical systems in neuroscience}.
\newblock MIT press, 2007.

\bibitem{gerstner2002spiking}
Wulfram Gerstner and Werner~M Kistler.
\newblock {\em Spiking neuron models: Single neurons, populations, plasticity}.
\newblock Cambridge university press, 2002.

\bibitem{gerstner1993spikes}
Wulfram Gerstner, Raphael Ritz, and J~Leo Van~Hemmen.
\newblock Why spikes? hebbian learning and retrieval of time-resolved
  excitation patterns.
\newblock {\em Biological cybernetics}, 69(5-6):503--515, 1993.

\bibitem{rumelhart1986learning}
David~E Rumelhart, Geoffrey~E Hinton, and Ronald~J Williams.
\newblock Learning representations by back-propagating errors.
\newblock {\em nature}, 323(6088):533--536, 1986.

\bibitem{lillicrap2020backpropagation}
Timothy~P Lillicrap, Adam Santoro, Luke Marris, Colin~J Akerman, and Geoffrey
  Hinton.
\newblock Backpropagation and the brain.
\newblock {\em Nature Reviews Neuroscience}, 21(6):335--346, 2020.

\bibitem{cho2022separable}
Junwoo Cho, Seungtae Nam, Hyunmo Yang, Seok-Bae Yun, Youngjoon Hong, and
  Eunbyung Park.
\newblock Separable pinn: Mitigating the curse of dimensionality in
  physics-informed neural networks.
\newblock {\em arXiv preprint arXiv:2211.08761}, 2022.

\bibitem{rueckauer2017conversion}
Bodo Rueckauer, Iulia-Alexandra Lungu, Yuhuang Hu, Michael Pfeiffer, and
  Shih-Chii Liu.
\newblock Conversion of continuous-valued deep networks to efficient
  event-driven networks for image classification.
\newblock {\em Frontiers in neuroscience}, 11:682, 2017.

\bibitem{li2021free}
Yuhang Li, Shikuang Deng, Xin Dong, Ruihao Gong, and Shi Gu.
\newblock A free lunch from ann: Towards efficient, accurate spiking neural
  networks calibration.
\newblock In {\em International Conference on Machine Learning}, pages
  6316--6325. PMLR, 2021.

\bibitem{liu2021optimal}
Yujie Liu, Zhe Wang, Peng Li, Yiran Chen, and Hai Li.
\newblock Optimal ann-snn conversion for fast and accurate inference in deep
  spiking neural networks.
\newblock {\em arXiv preprint arXiv:2105.11654}, 2021.

\bibitem{kim2021reducing}
Jihwan Kim, Jangho Lee, and Junmo Lee.
\newblock Reducing ann-snn conversion error through residual learning.
\newblock {\em arXiv preprint arXiv:2302.02091}, 2021.

\bibitem{snn_toolbox}
Bodo Rueckauer, Iulia-Alexandra Lungu, Yuhuang Hu, Michael Pfeiffer, and
  Shih-Chii Liu.
\newblock Spiking neural network conversion toolbox.
\newblock \url{https://github.com/NeuromorphicProcessorProject/snn_toolbox},
  2017.

\bibitem{li2022calibration}
Yuhang Li, Shikuang Deng, Xin Dong, and Shi Gu.
\newblock Converting artificial neural networks to spiking neural networks via
  parameter calibration, 2022.

\bibitem{liu2021backeisnn}
Zhen Liu, Zhihui Li, Zhihong Li, Yujie Wang, Zhen Li, Zhihui Li, Zhihong Li,
  and Yujie Wang.
\newblock Backeisnn: A deep spiking neural network with adaptive self-feedback
  and balanced excitatory–inhibitory neurons.
\newblock {\em Neural Networks}, 138:1–14, 2021.

\bibitem{jin2021nsfnets}
Xiaowei Jin, Shengze Cai, Hui Li, and George~Em Karniadakis.
\newblock Nsfnets (navier-stokes flow nets): Physics-informed neural networks
  for the incompressible navier-stokes equations.
\newblock {\em Journal of Computational Physics}, 426:109951, 2021.

\bibitem{ns_eqn1}
Xin-Yang Liu, Hao Sun, Min Zhu, Lu~Lu, and Jian-Xun Wang.
\newblock Predicting parametric spatiotemporal dynamics by multi-resolution pde
  structure-preserved deep learning.
\newblock {\em arXiv preprint arXiv:2205.03990}, 2022.

\bibitem{ns_eqn2}
Min Zhu, Handi Zhang, Anran Jiao, George~Em Karniadakis, and Lu~Lu.
\newblock Reliable extrapolation of deep neural operators informed by physics
  or sparse observations.
\newblock {\em Computer Methods in Applied Mechanics and Engineering},
  412:116064, 2023.

\bibitem{kim1985NSsolution}
John Kim and Parviz Moin.
\newblock Application of a fractional-step method to incompressible
  navier-stokes equations.
\newblock {\em Journal of computational physics}, 59(2):308--323, 1985.

\end{thebibliography}

\newpage
\section*{Appendix}
\begin{lemma}
    If the activation function is piecewise linear, then the conversion error in the final network output space can be bounded by a weighted sum of local conversion errors, given by
    \begin{equation}
        \e\att{n} \Hes\at{n} \e\at{n} \leq \sum_{l=1}^n 2^{n-l+1} \e_c\att{l} \Hes\at{l} \e_c\at{l}
    \end{equation}
    \label{eqn:pf:pwlinear}
    \end{lemma}
    \emph{Proof} In the proof of \cite{li2022calibration} uses only one property of ReLU that its second-order derivatives are all zeros. And it does not require the activation to be non-negative. So it applies to all piecewise linear activation functions.
    
    \begin{lemma}
    The Hessian matrix of activation in layer $l$ can be written as 
    \begin{equation}
        \Hes\at{l} = \W\att{l}\B\att{l}\Hes\at{l+1}\B\at{l}\W\at{l} + \W\att{l}\C\at{l}\W\at{l}
    \end{equation}
    where $W\at{l}$ is the weight of layer $l$,
    \begin{equation}
        \B\at{l} = \mathrm{diag}\left\{\frac{\partial\x\at{l+1}_1}{\partial\z\at{l}_1},\ldots,\frac{\partial\x\at{l+1}_{n_l}}{\partial\z\at{l}_{n_l}}\right\}
    \end{equation}
    is a diagonal matrix, each element on diagonal is the derivative of activation function, and
    \begin{equation}
        \C\at{l} = \mathrm{diag}\left\{\frac{\partial L}{\partial\x\at{l+1}_1}f''(z\at{l}_1),\ldots,\frac{\partial L}{\partial\x\at{l+1}_{n_l}}f''(z\at{l}_{n_l})\right\}
    \end{equation}
    here $x\at{l+1}$ is the input of layer $(l+1)$ and $z\at{l}$ is the unactivated output of layer $(l)$.
    \label{eqn:pf:Hes-expansion}
    \end{lemma}
    \emph{Proof} Similar to the proof in the \cite{li2022calibration}, we calculate $\Hes\at{l}_{a,b}$ by
    \begin{equation}
    \begin{aligned}
        \Hes\at{l}_{a,b} &= \frac{\partial^2 L}{\px\at{l}_a\px\at{l}_b} = \frac{\partial}{\px\at{l}_b}\left(\sum_i\frac{\pL}{\px\at{l+1}_i}\frac{\px\at{l+1}_i}{\px\at{l}_a}\right)\\
        &= \sum_i\frac{\partial}{\px\at{l}_b}\left(\frac{\pL}{\px\at{l+1}_i}\frac{\px\at{l+1}_i}{\pz\at{l}_i}\frac{\pz\at{l}_i}{\px\at{l}_a}\right)\\
        &= \sum_i\W\at{l}_{i,a}\frac{\partial}{\px\at{l}_b}\left(\frac{\pL}{\px\at{l+1}_i}\frac{\px\at{l+1}_i}{\pz\at{l}_i}\right)\\
        &= \sum_i\W\at{l}_{i,a}\frac{\px\at{l+1}_i}{\pz\at{l}_i}\frac{\partial}{\px\at{l}_b}\frac{\pL}{\px\at{l+1}_i} + \sum_i\W\at{l}_{i,a}\frac{\pL}{\px\at{l+1}_i}\frac{\partial}{\px\at{l}_b}\frac{\px\at{l+1}_i}{\pz\at{l}_i}\\
        &=\sum_{i,j}\W\at{l}_{i,a}\frac{\px\at{l+1}_i}{\pz\at{l}_i}\frac{\partial^2 L}{\px\at{l+1}_j\px\at{l+1}_i}\frac{\px\at{l+1}_j}{\pz\at{l}_j}\W\at{l}_{j,b}+\sum_{i}\W\at{l}_{i,a} \frac{\pL}{\px\at{l+1}_i}f''(\z_i)\W\at{l}_{i,b}
    \end{aligned}
    \end{equation}
    
    \begin{lemma}
        Let $M\in\mathbb{R}^{p\times p}$ and $x_n\in\mathbb{R}^p$. If $\norm{x_n}<K_x$ $\forall n$, $\norm{x_n-x_m}<\epsilon$, then $|x_n^{\top}Mx_n-x_m^{\top}Mx_m|<2K_MK_x\epsilon$, where $K_M = \norm{M} \stackrel{d}{=} \sup_{\norm{x}=\norm{y}=1}x^{\top}Ky$.
        \label{eqn:pf:blf-approx}
    \end{lemma}
    \emph{Proof} By the triangular inequality and definition of $K_M$,
    \begin{equation}
    \begin{aligned}
        |x_n^{\top}Mx_n-x_m^{\top}Mx_m|&\leq|x_n^{\top}Mx_n-x_m^{\top}Mx_n| + |\leq|x_m^{\top}Mx_n-x_m^{\top}Mx_m|\\
        &\leq K_M\norm{x_n}\norm{x_n-x_m}+K_M\norm{x_m}\norm{x_n-x_m}\\
        &< 2K_MK_x\epsilon
    \end{aligned}
    \end{equation}
    
    \begin{lemma}
        Let $x\in\mathbb{R}^p$. If $\norm{x}<\epsilon$, then $|x^{\top}\C\at{l}x|<K_l\sqrt{L}\epsilon$, where $K_l$ is a constant and $L$ is the training loss.
        \label{eqn:pf:C-bound}
    \end{lemma}
    \emph{Proof} Let $y = \W\at{l}x$, then $\exists K_W>0$, s.t. $\norm{y}\leq K_W\epsilon$. And since we are using MSE for the loss,
    \begin{equation}
        \frac{\pL}{\px\at{l+1}_i} = \sum_j\frac{\pL}{\px\at{n}_j}\frac{\px\at{n}_j}{\px\at{l+1}_i} \leq \sum_j K_J\sqrt{L}\frac{\px\at{n}_j}{\px\at{l+1}_i} \leq K\sqrt{L}
    \end{equation}
    for some $K_J, K>0$, because of the boundedness of $f'$ and $\W$. And $\norm{\C\at{n}}<\infty$ by the boundedness of $f''$, then by definition of $\norm{\C}$ we have the result.
    
    With the preceding lemmas, now we can prove our main result.
    
    \emph{Proof} Suppose the activation function is $f$. By the approximation property of $f$, $\forall \epsilon>0$, $\exists \tf$, such that
    \begin{equation}
        \norm{f-\tf}_\infty<\epsilon,\quad \norm{f'-\tf'}_\infty<\epsilon
        \label{eqn:pf:f-approx}
    \end{equation}
    Let $\Tilde{\cdot}$ be the values computed by the approximating activation function, like $\Tilde{\vb{x}}\at{l}$ be the output of layer $l$ computed by the approximating activation function $\Tilde{f}$. Define $\te$ as the error computed with $\Tilde{\vb{x}}$ instead of $\vb{x}$. Then we can assume
    \begin{equation}
        \norm{\te-\e}<\epsilon,\quad \norm{\tec-\ec}<\epsilon
        \label{eqn:pf:e-bound}
    \end{equation}
    by (\ref{eqn:pf:f-approx}).
    
    First we can approximate $\e\att{n}\tHes\at{n}\e\at{n}$ by $\te\att{n}\Hes\at{n}\te\at{n}$. This follows from Lemma \ref{eqn:pf:blf-approx} and the fact that $\tHes\at{n}=\Hes\at{n}$ as the last layer does not have an activation function to convert.
    
    Then we need to approximate $\ec\att{l}\Hes\at{l}\ec\at{l}$.
    \begin{equation}
        |\ec\att{l}\Hes\at{l}\ec\at{l} - \tec\att{l}\tHes\at{l}\tec\at{l}| \leq|\ec\att{l}\Hes\at{l}\ec\at{l} - \tec\att{l}\Hes\at{l}\tec\at{l}| + |\tec\att{l}\Hes\at{l}\tec\at{l} - \tec\att{l}\tHes\at{l}\tec\at{l}|
    \end{equation}
    The first term on the right hand side is easy to bound by Lemma \ref{eqn:pf:blf-approx}. To investigate the second term, we compute $\Hes\at{l}$ recursively from $l=n-1$. By Lemma \ref{eqn:pf:Hes-expansion},
    \begin{equation}
    \begin{aligned}
        \Hes\at{n-1} &= \W\att{n-1}\B\att{n-1}\Hes\at{n}\B\at{n-1}\W\at{n-1} + \W\att{n-1}\C\at{n-1}\W\at{n-1}\\
        &=\W\att{n-1}(\B\att{n-1}\Hes\at{n}\B\at{n-1}-\tB\att{n-1}\Hes\at{n}\tB\at{n-1})\W\at{n-1}\\
        &\quad+ \W\att{n-1}\tB\att{n-1}\Hes\at{n}\tB\at{n-1}\W\at{n-1} + \W\att{n-1}\C\at{n-1}\W\at{n-1}\\
        &= \W\att{n-1}\Delta\tB\at{n-1}\W\at{n-1} + \tHes\at{n-1} + \W\att{n-1}\C\at{n-1}\W\at{n-1}
    \end{aligned}
    \end{equation}
    where $\Delta\tB\at{n-1} = \B\att{n-1}\Hes\at{n}\B\at{n-1}-\tB\att{n-1}\Hes\at{n}\tB\at{n-1}$. Then by Lemma \ref{eqn:pf:blf-approx} and \ref{eqn:pf:C-bound}, we have $\Hes\at{n-1}-\tHes\at{n-1}\approx K_{n-1}\sqrt{L}\vb{I}$ in the sense of bilinear form. For $n=l-2$,
    \begin{equation}
    \begin{aligned}
        \Hes\at{n-2} &= \W\att{n-2}\B\att{n-2}\Hes\at{n-1}\B\at{n-2}\W\at{n-2} + \W\att{n-2}\C\at{n-2}\W\at{n-2}\\
        &= \tHes\at{n-2} + \W\att{n-2}\B\att{n-2}\Hes\at{n-1}\B\at{n-2}\W\at{n-2} + \W\att{n-2}\C\at{n-2}\W\at{n-2}\\
        &\quad- \W\att{n-2}\tB\att{n-2}\tHes\at{n-1}\tB\at{n-2}\W\at{n-2}\\
        &= \tHes\at{n-2} + \W\att{n-2}\Delta\tB\at{n-2}\W\at{n} + \W\att{n-2}\tB\att{n-2}\W\att{n-1}\Delta\tB\at{n-1}\W\at{n-1}\tB\at{n-2}\W\at{n-2}\\
        &\quad+ \W\att{n-2}\tB\att{n-2}\W\att{n-1}\C\at{n-1}\W\at{n-1}\tB\at{n-2}\W\at{n-2} + \W\att{n-2}\C\at{n-2}\W\at{n-2} 
    \end{aligned}
    \end{equation}
    Similarly, we have $\Hes\at{n-2}-\tHes\at{n-2}\approx K_{n-2}\sqrt{L}\vb{I}$ in the sense of bilinear form. Thus $\Hes\at{l}-\tHes\at{l}\approx K_{l}\sqrt{L}\vb{I}$ $\forall l=1,2,\ldots,n-1$ by induction. Hence we can approximate $\tec\att{l}\tHes\at{l}\tec\at{l}$ by $\ec\att{l}(\Hes\at{l} + K_l\sqrt{L}\vb{I})\ec\at{l}$. 
    
    Finally, because both sides of the inequality in Lemma \ref{eqn:pf:pwlinear} are well approximated, the proof is completed.

\end{document}